\documentclass[conference]{IEEEtran}
\usepackage{times}
\usepackage{multirow}
\usepackage[numbers]{natbib}
\usepackage{multicol}
\usepackage[bookmarks=true]{hyperref}
\usepackage{graphicx}
\usepackage{color}
\usepackage{xcolor}
\usepackage{array}
\usepackage{algorithm}
\usepackage{algorithmic}
\usepackage{amsmath}
\usepackage{amssymb}
\usepackage[justification=centering]{caption}

\usepackage[normalem]{ulem}
\usepackage{float}

\usepackage{graphicx}
\usepackage{subcaption}
\usepackage{xspace}
\usepackage{booktabs}
\usepackage{cleveref}
\let\labelindent\relax
\usepackage{enumitem}

\renewcommand{\baselinestretch}{0.97}
\newcolumntype{M}[1]{>{\centering\arraybackslash}p{#1}}
\usepackage[font=small,skip=8pt]{caption}
\newcolumntype{P}[1]{>{\centering\arraybackslash}p{#1}}
\newcommand{\method}[0]{\textit{URDFormer}\xspace}

\newcommand{\ie}[0]{\textit{i.e.}\xspace}
\newcommand{\eg}[0]{\textit{e.g.}\xspace}

\makeatletter
\patchcmd\@maketitle\null{{\titlefigure{}\par}}{}{}
\makeatother

\begin{document}

\title{URDFormer: A Pipeline for Constructing Articulated Simulation Environments from Real-World Images}

\author{Zoey Chen$^1$, Aaron Walsman$^1$, Marius Memmel$^1$, Kaichun Mo$^2$, Alex Fang$^1$, \\ Karthikeya Vemuri$^1$, Alan Wu$^1$, 
Dieter Fox*$^{1,2}$, Abhishek Gupta*$^{1,2}$
\smallskip 
\\
$^1$University of Washington ~~~
$^2$ Nvidia
\\[0.8em]
\textbf{\Large
\href{https://urdformer.github.io}{\color{blue}{urdformer.github.io}}}
}

\setcounter{figure}{0}

\makeatletter
\g@addto@macro\@maketitle{
    \begin{figure}[H]
  \setlength{\linewidth}{\textwidth}
  \setlength{\hsize}{\textwidth}
  \centering
  \resizebox{1.\textwidth}{!}{
  \includegraphics[]{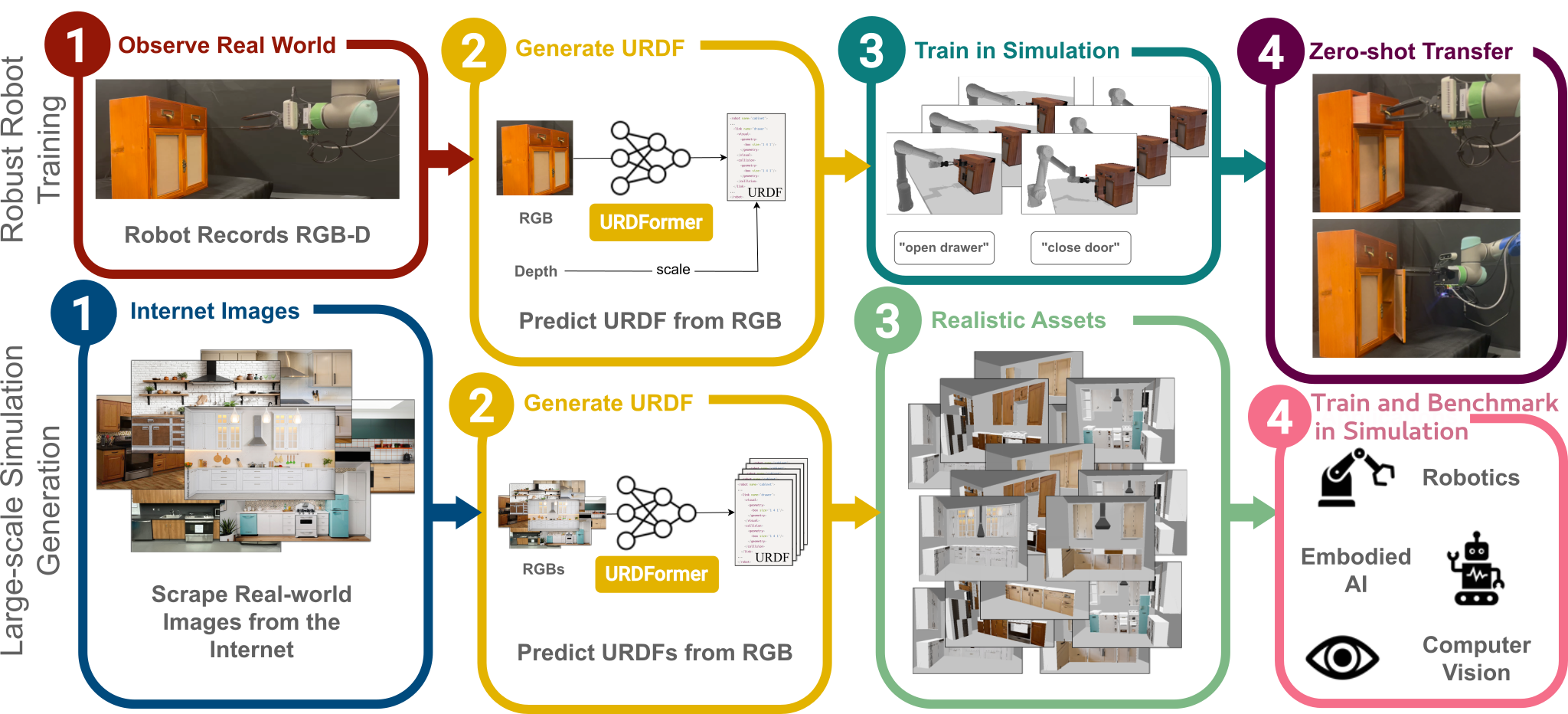}
  }
  \caption{\footnotesize{The \method predicts realistic, kinematic scenes from images that 1) allow for zero-shot real-to-sim-to-real transfer through targeted randomization, and 2) generate internet-scale simulation assets valuable to a multitude of applications.}}
  \label{fig:main}
  \end{figure}
  \vspace{-6mm}
}
\makeatother

\maketitle

\setcounter{figure}{1}

\begin{abstract}
Constructing accurate and targeted simulation scenes that are both visually and physically realistic is a problem of significant practical interest in domains ranging from robotics to computer vision.
This problem has become even more relevant as researchers wielding large data-hungry learning methods seek new sources of training data for physical decision-making systems.
However, building simulation models is often still done by hand - a graphic designer and a simulation engineer work with predefined assets to construct rich scenes with realistic dynamic and kinematic properties. While this may scale to small numbers of scenes, to achieve the generalization properties that are required for data-driven robotic control, we require a pipeline that is able to synthesize large numbers of realistic scenes, complete with ``natural" kinematic and dynamic structure. To attack this problem, we develop models for inferring structure and generating simulation scenes from natural images, allowing for scalable scene generation from web-scale datasets. To train these image-to-simulation models, we show how controllable text-to-image generative models can be used in generating paired training data that allows for modeling of the \emph{inverse} problem, mapping from realistic images back to complete scene models. We show how this paradigm allows us to build large datasets of scenes in simulation with semantic and physical realism. We present an integrated end-to-end pipeline that generates simulation scenes complete with articulated kinematic and dynamic structures from real-world images and use these for training robotic control policies. We then robustly deploy in the real world for tasks like articulated object manipulation. In doing so, our work provides both a pipeline for large-scale generation of simulation environments and an integrated system for training robust robotic control policies in the resulting environments.  

\end{abstract}


\IEEEpeerreviewmaketitle
\section{Introduction}
Simulation has become a cornerstone of a plethora of applied machine learning problems - from the natural sciences such as physics, chemistry, and biology ~\citep{jia21pgml, mlmodel} to robotics ~\citep{collins21physics, narang22factory} and computer vision~\citep{muller18sim4cv, memmel2023modality}. Simulation allows for scalable and cheap data collection while providing an easy way to encode domain-specific prior knowledge into end-to-end machine learning problems. This is particularly important for data-scarce problems such as robotics, where collecting real data can lead to costly and unsafe failures or may require expensive human supervision. Critical to each of these endeavors is a rich and accurate simulation environment, complete with assets depicting complex scene layouts and kinematic structure. For instance, advances in robotic mobile manipulation in the Habitat simulator~\citep{szot21habitat}, are critically dependent on the Matterport dataset for realistic scenes ~\citep{yadav23habitatmatterport}. The creation and curation of these simulation scenes and assets is an important but often overlooked part of the process. 

The most common approaches for generating simulation content are either manual ~\citep{kolve2017ai2, gupta2023predicting}, procedural ~\citep{deitke2022} or, more recently, purely generative~\cite{dreamfusion, wang2023robogen, wang2023gen, liu2023one, fang2023ctrl}.
The manual process for creating simulation scenes requires a designer to characterize, identify, and model a particular real-world scene, a painstaking and impractical process. While this approach can produce high quality results, it often leads to content that lacks diversity due to the amount of human effort required. On the other hand, rule-based procedural generation methods~\citep{deitke2022, raistrick23procedural} have been applied in robotics applications such as navigation, but often struggle to capture the natural complexity of the real world for problems such as manipulation. Moreover, the procedural generation process is not \emph{controllable}, making it hard to generate simulation content corresponding to a \emph{particular} real-world environment, which is often important in real-world robotic learning pipelines. Recently introduced generative methods for content creation~\cite{dreamfusion, liu2023one, fang2023ctrl, wang2023robogen} can generate visually appealing 3-D geometries for particular objects, but often result in  undesired physical simulation behavior and lack kinematic structure like articulation. 

What are the desiderata for a content creation method for simulation? To enable a variety of downstream use cases such as robotic learning, scalable content creation in simulation must be (1) realistic enough such that machine learning models trained in the constructed simulation environments transfer back to the real world, (2) diverse in a way that captures the statistics of natural environments so as to enable learning generalizable models and policies (3) controllable in a way that allows for targeted generation of particular scenes of interest. 

To generate content of this nature, we develop a pipeline to train a transformer-based network, \method, that maps directly from individual real-world images to corresponding simulation content (expressed as a Unified Robot Description File (URDF)) that could plausibly represent the semantics, kinematics, and structure of the scene (Fig\ref{fig:main}).
We leverage controllable text-to-image generative models~\citep{rombach2022high} to generate a large-scale paired dataset of structured simulation scenes and closely corresponding, realistic images.  This paired dataset is then \emph{inverted} to train \method, which maps from RGB images directly to plausible simulation environments with semantic and kinematic structure. 

\method can then naturally be used in several use cases --- (1) diverse content generation for simulation: generating a large and diverse set of realistic simulation environments that correspond directly to uncurated, real-world RGB images (e.g scraped off the web), or (2) targeted generation: generating a simulation environment (or narrow distribution of environments) corresponding to a particular set of desired images. We show that the generated simulation environments are a useful tool for robotic learning in a real-to-sim-to-real pipeline for training robotic control policies. 
%
%

\section{URDFormer - A Pipeline for Scalable Content Creation for Simulation from Real-World Images}
\label{sec:method}

Generating simulated scenes with a high degree of visual realism that supports rich kinematic and dynamic structure, while reflecting the natural statistics of the real world is a challenging problem.
Downstream applications in robotics and computer vision typically require data that is simultaneously \textit{realistic}, \textit{diverse}, and \textit{controllable}.

To accomplish these requirements, we attempt to generate rich kinematic scenes directly from RGB images, so that we can use either a large web-scale dataset of images harvested from the internet to generate a large number of realistic scenes, or a small number of domain-specific images to train robot models for more specific targeted environments.  Unfortunately, no large dataset containing images along with corresponding articulated scene descriptions is publicly available to train such a model.  We address this by building a new dataset using generative image models to enhance the visual quality of synthetic renders of procedurally sampled scenes.  This process requires us to reason about two distinct pipelines.  The \textit{forward} pipeline takes a procedurally sampled environment, renders it, and augments the resulting images with a generative model to produce a dataset of scene/image pairs.  The second \textit{inverse} pipeline then trains a model which takes in the images produced in the forward pipeline and produces the kinematic scene description that was used to create it.

Once the inverse model is trained, it allows for scene generation that is \textit{realistic}
since it can be used to build scenes that correspond to real images.
The generated scenes are \textit{diverse} since large web-scale image datasets with diverse content can be used to seed this generation process. Lastly, the generation is \textit{controllable} since curated images of particular target environments can be used to generate corresponding simulation assets. We first define the inverse problem of synthetic scene generation from real-world images, then describe how to learn inverse models to solve this problem with supervised learning on a paired dataset generated using controllable text-to-image generative models~\cite{rombach2022high}. Finally, we show how the learned inverse model can be used with real-world image datasets for scalable content creation. This tool can then be used to instantiate a real-to-simulation-real pipeline for robot learning, as described in Section~\ref{sec:pipeline_real2sim2real}. 


\begin{figure}[!h]
    \centering
    \includegraphics[width=0.49\textwidth]{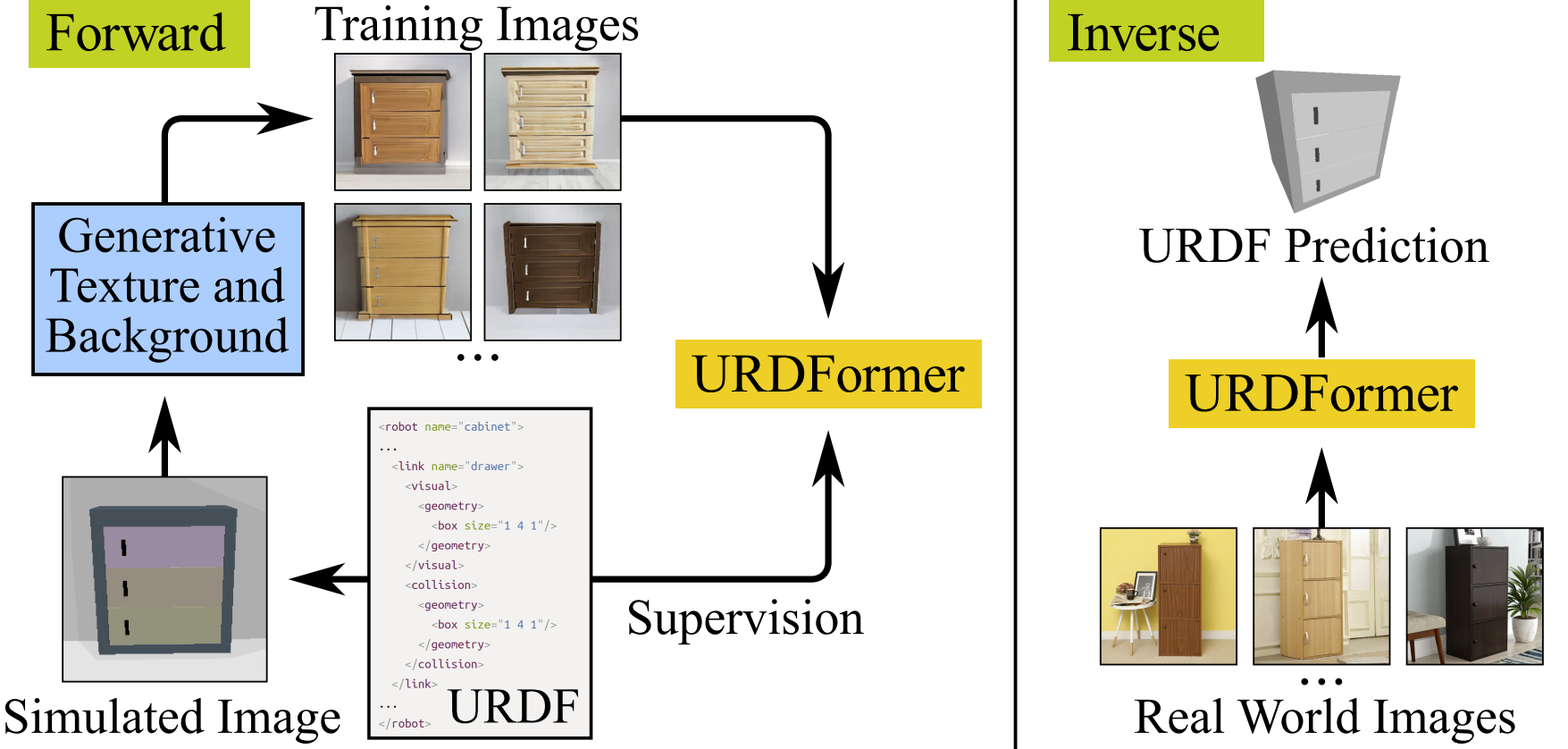}
    \caption{The \method is trained on a large paired dataset of simulation assets and realistic renderings (forward). During inference, this process is inverted and it predicts the URDF from a real image (inverse).}
    \vspace{-1.5em}
    \label{fig:forward_inverse}
\end{figure}

\begin{figure*}[!t]
    \centering
\includegraphics[width=0.98\textwidth]{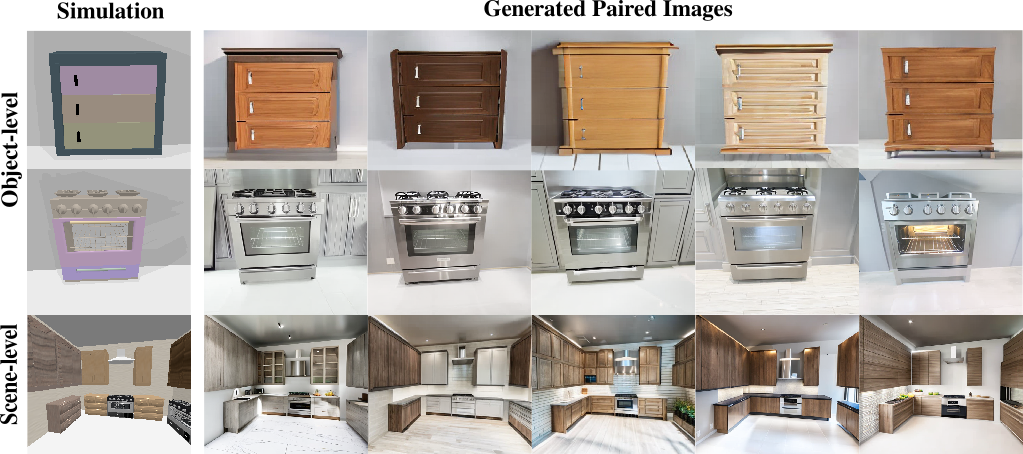}
    \caption{\textbf{Controlled Generation:} Rendering URDF models in simulation and generating paired images with a guided diffusion model.}
    \vspace{-1.3em}
    \label{fig:data_generation}
\end{figure*}
\subsection{Problem Formulation}
\label{sec:problem_formulation}
To formalize the problem of simulation scene generation from real-world images, let us consider a kinematic scene description $z$ drawn from a target scene distribution $P(z)$ in the real world. For our purposes, the scene can be described as a list of objects $z = \left\{ o_1 \hdots o_n\right\}$, where each object $o_i$ contains a base class label $c_i$, a 3D bounding box $b_i \in \mathbb{R}^6$, a 3D transform $T_i \in SE(3)$, a kinematic parent $p_i \in [1 \hdots i-1]$ and a joint type $j_i$ that specifies how that object can move relative to its parent $o_i = \left(c_i, b_i, T_i, p_i, j_i\right)$. This resembles the typical representation of scenes and robots using the unified robot description format (URDF). 
Let's consider a kitchen scenario that contains a row of cabinets next to a stove. The cabinets and the stove will be the kinematic children of the wall, and the kinematic parents of their respective doors. Similarly, these doors will be the kinematic parents of their handles. 
As the example shows, the kinematic structure $z$ for a particular real-world scenario is unknown without extensive human labeling effort, and instead, we only have access to the result $x$ of an indirect ``forward'' function $f$, $x = f(z)$. For example, $x$ could be a photograph of the real environment, or a point cloud captured with a LIDAR scanner. The goal of this work is to recover the entire kinematic and semantic structure of the scene from just having access to the forward evaluation $x$, requiring complete inference of a rich scene representation $z$. 



Unfortunately, since the scene structure $z$ is unknown for most complex real-world scenes and difficult to generate manually, it is challenging to solve the ``inverse'' generation problem to infer the scene description $z$ from the forward rendered images (or alternative sensor readings) $x$, $z = f^{-1}(x)$. Had there been a sizeable dataset $\mathcal{D} = \{(z_i, x_i)\}_{i=1}^N$ of scene descriptors $z_i$ in simulation and their corresponding real-world counterparts $x_i$, the inverse problem could be solved using supervised learning (minimizing a loss $\mathcal{L}$ like the cross entropy loss or a MSE loss) to learn an $f^{-1}_\theta$ that approximates the scene structure $\hat{z}$ given an input forward-rendered image $x$.

However, such a paired dataset does not readily exist, making direct application of supervised learning methods challenging. In this work we take an inversion through synthesis approach\textemdash leveraging pre-trained generative models to convert procedurally generated scenes in simulation into a large paired dataset of scene content $z$ and corresponding realistic RGB images $x$.  This process can generate a large and diverse dataset of image and scene-description $(x,z)$ pairs that we can use to train an approximate inverse model $f^{-1}_\theta(x)$ that generates scene descriptions $\hat{z}$ from real RGB images $x$. Since most scenes that we consider are object-centric, we decompose the inverse problem into two parts: (1) object-level prediction that focuses on the kinematic structure of individual objects, \eg, cabinets, and (2) global-scene prediction, \eg, kitchens, that focuses on the structure of an overall scene. We next discuss the process of generating a large paired dataset for these two components (Section~\ref{sec:paireddatasetgen}) and then show the training process for the inverse model in detail (Section~\ref{sec:traininverse}).

\subsection{Controlled Generation of Paired Datasets with Pretrained Generative Models}
\label{sec:paireddatasetgen}

Given a simulated scene $z$ (drawn from a dataset such as PartNet~\citep{mopartnet}, or procedurally generated), we use the fact that controllable generative models~\cite{rombach2022high} are both diverse and realistic enough to take an unrealistic simulation rendering of a scene and generate a distribution of corresponding \emph{realistic} images. This allows the scene in simulation with unrealistic appearance and texture to be translated into a diverse set of visually realistic images that plausibly match the same underlying environment, as shown on the left side of Figure \ref{fig:forward_inverse}. To ensure piecewise consistency and realism of the generated images, we use two different dataset generation techniques for the global scene structure and local object structure respectively. These share the same conceptual ideas but differ to account for consistency properties in each case. 
\begin{figure*}[t]
    \centering
    \includegraphics[width=\textwidth]{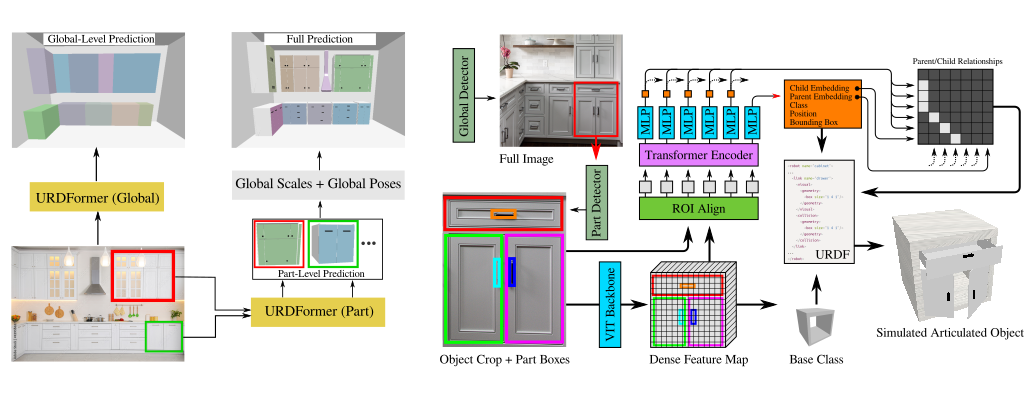}
    \vspace{-3.7em}
    \caption{Depiction of the \method Training Procedure and Architecture. (\textbf{Left}) Given an RGB image of the scene, \ie a kitchen, we train two separate networks: URDFormer (Global) focuses on predicting parent and spatial info of how to place the object. URDFormer (Part) takes the cropped image containing each object and predicts detailed structure. The results of the two predictions are combined and create the full scene prediction. (\textbf{Right}) The \method architecture takes as input a cropped RGB image and object part boxes and predicts a hierarchy consisting of a base class and parent-child relations that make up the final URDF file.}
    \vspace{-1.5em}
    \label{fig:urdformer_global}
\end{figure*}

\textbf{Scene-Level Dataset Generation:} To generate training data for the scene model, we feed a poorly rendered image from simulation along with a templated text prompt to an image-and-text guided diffusion model~\citep{rombach2022high}, as shown in Fig~\ref{fig:data_generation}. This generates a new image that attempts to simultaneously match the content described in the text prompt while retaining the global scene layout from the provided image.  We found that this model is able to reliably maintain the scene layout, but it may change some individual components of the scene, \eg, replacing objects with a different but plausible category, or changing the number of subcomponents within an object such as the drawers or handles. Despite these failures, the large-scale structural consistency still provides a useful source of training data. After running our simulated image through the generative model, we have realistic images that contain known high-level object positions and spatial relationships, but unknown category and low-level part structures (\eg the parts for articulated objects such as cabinets), since these may have been modified by the generative model in it's forward pass. This means that the scene model dataset contains complete images, but incomplete labels. Rather than complete $(x,z)$ pairs, we have a dataset $\mathcal{D}_{\text{scene}} = \{(x, \tilde{z})\}$ of $(x, \tilde{z})$ pairs where $\tilde{z}$ only contains the bounding boxes, transforms and parents of the high-level (non-part) objects $\tilde{z} = \left\{(b_1, T_1, p_1) \hdots (b_n, T_n, p_n) \right\}$ but lacks accurate low-level information.

\textbf{Object-Level Dataset Generation:} The process for generating object-level training data is similar but requires more care due to the tendency of text-to-image generative models to modify low-level details. For objects with complex kinematic structures, such as cabinets, we procedurally generate a large number of examples of these objects and render them in isolation from different angles.  Rather than using a generative model to construct entirely new images, we use it to produce diverse texture images, which are overlaid in the appropriate locations on the image using perspective warping. We then change the background of the image using the generative model with appropriate masking derived from the original rendering. This part-by-part texture-based rendering process ensures diversity while maintaining consistency in low-level details. Unlike the scene dataset which contains complete images but partial labels, the object dataset contains partial images (in the sense that they contain only a single object), but complete labels for the object and its kinematic parts. We can say that this dataset $\mathcal{D}_{\text{object}}$ contains $(\tilde{x}, z)$ pairs where $\tilde{x}$ is an image of a single object rather than a full scene (hence the partial $x$), and $z$ is complete for the single object and its parts. 

The result of these two data generation processes is a high-level scene structure dataset $\mathcal{D}_{\text{scene}}$, and a low-level object dataset $\mathcal{D}_{\text{object}}$, that can subsequently be used to train an object-level and a scene-level inverse model, as shown on right side in Figure \ref{fig:forward_inverse}.

\subsection{Learning Inverse Generative Models for Scene Synthesis}
\label{sec:traininverse}
Given the datasets $\mathcal{D}_{\text{object}} = (\tilde{x}, z)$ and $\mathcal{D}_{\text{scene}} = (x,\tilde{z})$ constructed as described above, we can use supervised learning methods to learn an \emph{inverse model} that maps images of a complex object or scene to the corresponding simulation asset. To take advantage of these partially complete datasets, we must add some structure to our prediction model. We do this by splitting our learned inverse model in correspondence with the split in our forward data generation process: we train one network $f^{-1}_\theta$ to predict the high-level scene structure using dataset $\mathcal{D}_{\text{scene}}$ and another network $g^{-1}_\phi$ to predict the low-level part structure of objects using $\mathcal{D}_{\text{object}}$.

To model both the scene-level prediction model ($f^{-1}_\theta$) and the low-level part prediction model ($g^{-1}_\phi$), we propose a novel network architecture --- \method, that takes an RGB image and predicts URDF primitives as shown in Fig\ref{fig:urdformer_global}. Note that both the scene-level prediction and the low-level part prediction use the same network architecture, the scene-level simply operates on full images with
object bounding boxes extracted, while the part-level operates on crops of particular objects with
object parts extracted. In the \method architecture, the image is first fed into a vision transformer~\citep{dosovitskiy2020image} (ViT) visual backbone to extract global features. We then obtain bounding boxes of the objects in the image using the masks rendered from the original procedurally generated scene in simulation (these are known at training time, and can be extracted using
detection
models at test time).  We then use ROI alignment~\citep{he2017mask} to extract features for each of these bounding boxes. These feature maps are combined with an embedding of the bounding box coordinates and then fed through a Transformer~\citep{vaswani2017attention} to produce a feature for each object in the scene. An MLP then decodes these features into an optional base class label (used only when training the object-level model), and a discretized 3D position and bounding box. In addition, it also produces a child embedding and a parent embedding that are used to predict the hierarchical relationships in the scene (object to its parent and so on). To construct these relationships, the network uses a technique from scene graph generation~\citep{yang2023pvsg} that produces an $n \times n$ relationship score matrix by computing the dot product of every possible parent with every possible child.
The scene-level model also specially has a set of learned embeddings for six different root objects corresponding to the four walls, the floor, and the ceiling so that large objects like countertops and sinks can be attached to the room.

\textbf{Test-time scene generation from real-world RGB images:} Due to the unpredictable nature of the text-to-image generative transforms that are used to perform global dataset generation, which may change the base class identities, \eg, the diffusion model might turn a fridge into a cabinet, only the position, bounding box, and relationship information is used when computing the high-level scene structure. 
To generate a full approximation of the scene structure $\hat{z}$ from a natural image at test time, the image and a list of high-level bounding boxes
(from a detection model)
are first fed into the scene prediction model $f^{-1}_\theta$, which predicts the global structure, i.e. the location and parent for each object.  The image regions corresponding to these boxes are then extracted and
a second detection model is used
to produce part-level bounding boxes. Each of these image regions that correspond to a particular object (e.g. a cabinet or a fridge) and the corresponding part-level boxes (e.g. individual drawers or doors) are then fed into the part prediction model $g^{-1}_\phi$ to compute the kinematic structure of the low-level objects and parts. This nested prediction structure can be used to generate entire scenes from web-scraped RGB images drawn from any image dataset to generate novel simulation content both at the scene level and at the object level. We visualize this process in Fig \ref{fig:urdformer_global}




\section{Using \method for robotic control via a real-to-simulation-to-real pipeline}
\label{sec:pipeline_real2sim2real}
As described above, \method provides the ability to generate realistic, diverse and controllable scenes in simulation through inverse modeling. This suggests that on deployment, \method allows a system to take a single picture of a deployment scene and then construct a fully articulated scene in simulation with \emph{minimal} human effort. In this section, we describe how this type of controllable inverse modeling can serve as a useful tool for robotic learning through a real-to-simulation-to-real pipeline.  

The most straightforward way of using \method for robotic control is a model-based one - first synthesize a ``digital twin" for a deployment time scene, and directly use this for planning and control in the real world with a model-based approach. This is challenging for several reasons - (1) the constructed simulations may not be perfectly accurate, (2) there is no access to Lagrangian environment state in the real world. Instead, we take a learning-based approach to the problem; we use \method to generate not precisely the test time scene in simulation, but rather a narrow, representative distribution of simulation environments via a targeted randomization procedure. This distribution of environments in simulation can be used to learn generalizable robotic policies that operate from raw perceptual input, directly transferring back from simulation to the real-world. Doing so closes the real-to-simulation-to-real loop, obtaining real-world robotic policies with minimal human effort in the process. This pipeline has three major components:

\textbf{Scene Generation:} Given a robot's RGB pointcloud observation of an unseen environment, we use \method to generate a URDF file from RGB that captures the kinematic and dynamic structure of the real-world scene. We further resize the URDF to fit the pointcloud's scale. Importing the URDF into simulation then provides a playground for data collection and policy training. Most importantly, this simulation is not just an arbitrary model but an approximate representation of the real world scene of interest on deployment.

\textbf{Targeted Randomization:} In simulation, we have access to ground truth information which we can exploit to inexpensively collect trajectory data for policy learning. We use an efficient motion planner~\cite{curobo_report23} to quickly collect approximately optimal trajectories solving multiple tasks in simulation. To increase the simulation diversity and decrease the sim2real gap, we additionally randomize over minor details, \eg, texture variants, shapes, and sizes of parts. This ``targeted" randomization is in contrast to procedural generation which covers many different environment variations but is not informed by the real-world environment.

\textbf{Policy Synthesis:} To synthesize a policy from the collected data, we can train a language-conditioned behavior cloning policy~\cite{yuan2023m2t2} operating from RGB point-clouds in simulation, applying image augmentations during training to enable policy transfer back to the real world. We stress that the proposed pipeline is not limited to data collection with motion planning and behavior cloning but can also be used to train policies with other policy search methods~\cite{memmel2022dimensionality, schulman2017proximal, haarnojasac}. 
\begin{figure*}[t]
    \centering
    \includegraphics[width=\textwidth]{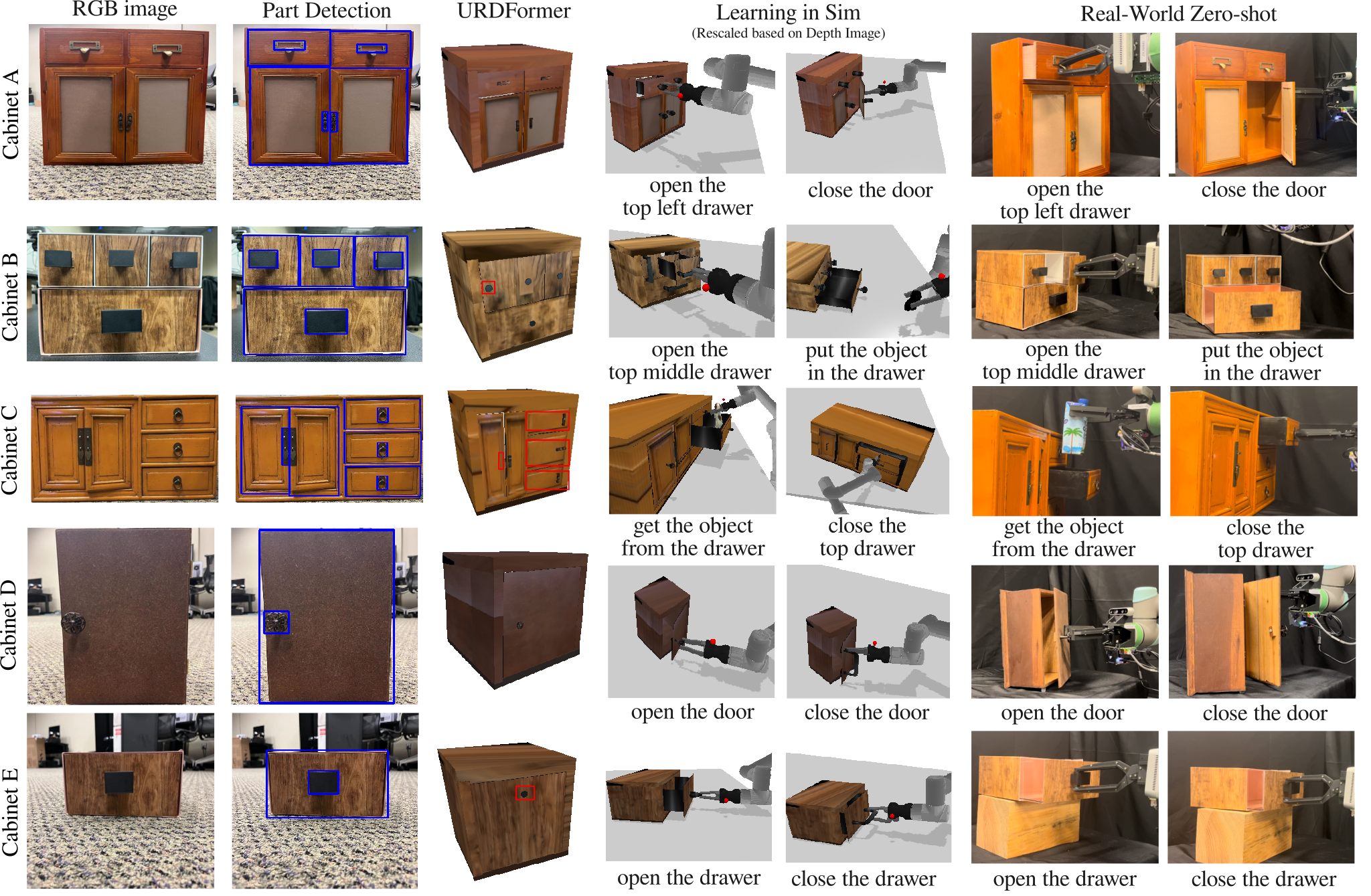}
    \caption{\textbf{Qualitative Results for Real-world Robot Experiments:} An RGB image is pre-processed by detecting bounding boxes of relevant parts. The \method then predicts the corresponding URDF of the cabinet. When importing the cabinet into the simulation, it is re-scaled using depth measurements. Furthermore, the real-world texture is cropped using the bounding boxes and projected onto the cabinet. This realistic simulation can then be used to generate massive data with the help of motion planning, ground truth information, and targeted domain randomization. Finally, we show that training a language-conditioned multi-task policy can be zero-shot transferred to the real world to solve several opening and closing tasks.}
    \vspace{-1.4em}
    \label{fig:real_world}
\end{figure*}
\begin{table*}[tb]
\centering
\begin{tabular}{M{2.2cm}M{2.2cm}M{2cm}M{2cm}M{2cm}M{2cm}M{2cm}}
                     & \multicolumn{2}{c}{\textbf{Cabinet A}}                & \multicolumn{2}{c}{\textbf{Cabinet B}}                           & \multicolumn{2}{c}{\textbf{Cabinet C}}                            \\
\toprule
Task                 & Open the top left drawer & close the door          & put object in bottom drawer & open top middle drawer & get object from middle drawer & close the top drawer \\
\midrule
OWL-ViT~\cite{minderer2022simple}              & 0/5                     & 0/5                     & 0/5                     & 0/5                     & 0/5              & 0/5              \\
DR    & 0/5                     & 0/5                     & 0/5                     & 0/5                     & 0/5              & 0/5              \\
URDFormer-ICP        & 2/5                     & 1/5                     & ----                    & 3/5                     & 1/5              & \textbf{3/5}              \\
URDFormer-TR         & \textbf{4/5}                     & \textbf{5/5}                     & \textbf{2/5}                     & \textbf{4/5}                     & \textbf{3/5}              & \textbf{3/5}                         \\
\bottomrule
\\

                     & \multicolumn{2}{c}{\textbf{Cabinet D}}                & \multicolumn{2}{c}{\textbf{Cabinet E}}                           & \multicolumn{2}{c}{\textbf{Average}}                             \\
\toprule
Task                 & Open the door            & close the door          & open the drawer             & close the drawer            & \multicolumn{2}{c}{}                \\
\midrule
OWL-ViT~\cite{minderer2022simple}              & 0/5                  & 0/5                 & 0/5                     & 0/5                     & \multicolumn{2}{c}{0/50}                \\
DR     & 0/5                  & 2/5                 & 2/5                     & \textbf{5/5}                     & \multicolumn{2}{c}{9/50}                \\
URDFormer-ICP        & 0/5                  & 3/5                 & 3/5                     & 2/5                     & \multicolumn{2}{c}{24/45}                \\
URDFormer-TR         & \textbf{5/5}                  & \textbf{4/5}                 & \textbf{4/5}                     & \textbf{5/5}                     & \multicolumn{2}{c}{\textbf{39/50}}                \\
\bottomrule
\end{tabular}
\caption{\textbf{Quantitative Results for our Real-world Robot Experiments:} Our Real2Sim2Real pipeline with the \method and targeted (domain) randomization in simulation results in a 78\% success rate across all tasks. Results are reported in success / number of trials.}
\label{tab:real_world_robot}
\vspace{-1.7em}
\end{table*}

The proposed pipeline results in a robust policy that can successfully solve tasks in the real-world without the manual burden of constructing environments in simulation~\cite{deitke2022}, expensive human data collection~\cite{open_x_embodiment_rt_x_2023} or real-world reinforcement learning~\cite{gupta21mtrf}. 
 
\section{Experiments}
\label{sec:experiment}

In this section, we aim to answer the following questions: 
\begin{enumerate}[label=\Alph*.]
    \item Does integrating \method in a real2sim2real pipeline improve policies that can transfer zero-shot to the real world?
    \item Can \method generate plausible and accurate simulation content from internet images?
    \item Can \method generalize to diverse objects and scenes?
    \item
    Can \method support different robots and tasks?
    
\end{enumerate}
\subsection{Does integrating \method in a real2sim2real pipeline improve policies that can transfer zero-shot to the real world?}
\label{sec:two_a}
As described in Section~\ref{sec:pipeline_real2sim2real}, \method can be used to instantiate a real-to-simulation-to-real pipeline for robot learning. In this section, we describe a concrete instantiation of one such pipeline and provide a detailed evaluation of the resulting robotic behavior in the real world. 

\textbf{Real-to-sim-to-real pipeline:}
We implement our real-to-sim-to-real approach on a UR5 robot equipped with a custom-made 3D printed 2-fingered gripper and an Intel RealSense D435i mounted on the endeffector. We evaluate our pipeline on five different cabinets with varying sizes, shapes, textures, joint types (revolute and prismatic), and handles, with two tasks per cabinet. The details of each pipeline element is as follows:
\newline
\noindent
\textbf{1. Real-to-sim:} First, our system takes an image of the scene and uses a finetuned Grounding DINO~\cite{liu2023grounding} with model soup approach\cite{wortsman2022model} (Detailed in \ref{sec:Grounding_dino_appendix}) to automatically detect the parts of the cabinet, \ie, door, drawer, and handles. \method then generates the corresponding URDF from the RGB image and the predicted bounding boxes using \method, as described in Section~\ref{sec:traininverse}. 
\newline
\noindent
\textbf{2. Policy Learning in Sim:} The URDF is then imported into a physics simulator (PyBullet~\cite{coumans2021}) and scaled appropriately using depth measurement. Next, a motion planner, in this case cuRobo~\cite{curobo_report23}, generates trajectories that solve a variety of tasks, \eg, closing/opening a drawer/door by utilizing privileged information in simulation. To account for possible prediction inaccuracies of \method (red highlights in Fig \ref{fig:real_world}) and to robustify the trained policy, we apply targeted randomization (TR) that randomizes the scene while maintaining its semantic configurations. In particular, doors, drawers, and handles are randomly replaced with their PartNet~\cite{mopartnet} equivalents while the size and base of the object are kept fixed.  For each door or drawer that was generated in this process, the geometry was randomly replaced with alternate geometry from the same PartNet class, but rescaled to be the appropriate size.  Additionally each handle or knob that was generated was similarly replaced with alternate geometry, but also randomly translated in the plane of the door or drawer that it was attached to. Textures are randomized by cropping out the real texture using the bounding boxes, generating variations by prompting Stable Diffusion~\citep{rombach2022high} and fitting it back onto the shape. Finally, the RGB input is augmented by adding standard augmentations such as Gaussian noise and color jitter.
After automatically collecting a dataset of successful simulation trajectories using the motion planner, we train a behavior cloning policy network that predicts end-effector poses from point clouds. The network architecture follows M2T2~\cite{yuan2023m2t2} and predicts 6D end-effector poses from RGB pointclouds given language instructions specifying the task. Appendix D-C provides the full training procedure and architectural details.
\newline
\noindent
\textbf{3. Sim-to-Real:} In order to transfer back to the real world, the policy takes an RGB pointcloud, current end-effector pose and a natural language instruction, and predicts the next end-effector pose, using a PD controller to execute these predictions. 

We term this specific instantiation of our real2sim2real pipeline \method-TR and stress that the data generation process, the policy network, and the transfer procedure are highly flexible and can be replaced depending on the exact problem setting.
 
\textbf{Baselines:} We compare \method-TR with multiple variations of our real-to-simulation-real pipeline with varying degrees of access to real-world information.  The results are shown in Table \ref{tab:real_world_robot}.

(1) \textbf{OWL-ViT~\cite{minderer2022simple}:} First, we evaluate against a zero-shot vision-language model baseline. Inspired by VoxPoser \cite{huang2023voxposer}, given a language instruction (Appendix D-B), we used the same open-vocabulary detector (OWL-ViT~\cite{minderer2022simple}) to predict bounding boxes for the prompted parts and handles that are important to the task. After mapping the detection to the observed pointcloud, a motion planner can then generate plans to solve the task directly in the real world.

(2) \textbf{Domain Randomization (DR)~\cite{tobin17domainrand}:} Another approach is Domain Randomization (DR). We follow the augmentations from \method-TR but on randomly generated cabinets with different configurations. Similarly to \method-TR, since we assume depth observations during inference, we also scale the generated cabinets to the real-world size. We follow the same approach for trajectory generation, policy training, and real-world transfer procedure of \method-TR.

(3) \textbf{\method-ICP:} This presents a learning-free, digital twin-style approach based on the Iterative Closest Point (ICP)~\cite{besl1992method} algorithm. First, we construct a simulation with the URDF created by the \method and scale it according to the depth observation. We then use the ground truth simulation to compute end effector poses which we execute directly in the real world. When the cabinet's pose changes, we use ICP to transform the computed end effector poses to the new cabinet pose.

\textbf{Real-world Results of Real-to-Sim-Real Training:}
Running the zero-shot OWL-ViT, we find that the model fails to predict the fine-grained details required to solve the tasks, \ie, "top middle drawer", "right door", and "handles".
We visualize qualitative examples of results from OWL-ViT in Appendix D. Without localizing these regions of interest, the motion planning cannot solve the tasks leading to 0\% success on all tasks.

While DR works surprisingly well on simple cabinets, \eg, Cabinet E which only has a single drawer, it fails to solve any of the more complicated configurations. We observe \method-TR outperforms DR by 40\% on average, showing the benefits of targeting the domain randomization procedure to the real-world configuration.

\method-ICP shows average performance across all opening and closing tasks. When tasking it with ''put object in bottom drawer'', we observe the limitations of the approach. Since the object is a lot smaller than the cabinet, ICP matches the pointcloud of the cabinet instead of the object. This results in an inability to transform the endeffector pose with respect to the graspable object and results in failure to solve the task. In general, neither baseline can reliably solve putting and getting objects in/from the drawer. While \method-TR succeeds 50\% of the time, it showcases the difficulty of the task and leaves space for future improvements.
\begin{figure*}[t]
    \centering
    \includegraphics[width=\textwidth]{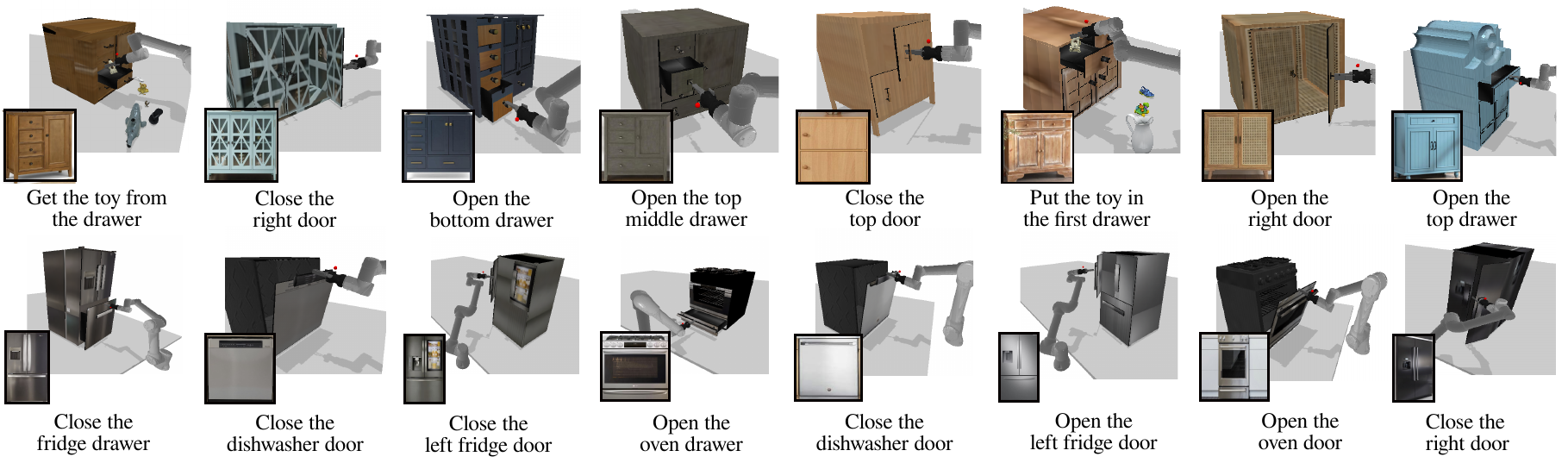}
    \caption{\textbf{Reality Gym:} A simulation environment with a variety of assets originated from internet images (black box) using \method. We predict URDFs of internet images which can be loaded in any simulator. These URDFs are randomized with meshes from the Partnet dataset. We introduce 4 main tasks: (1) Open any articulated parts (2) close any articulated parts (3) fetch objects and (4) collect objects}
    \label{fig:reality_gym_suite}
\end{figure*}

\begin{figure*}[t]
    \centering
    \includegraphics[width=\textwidth]{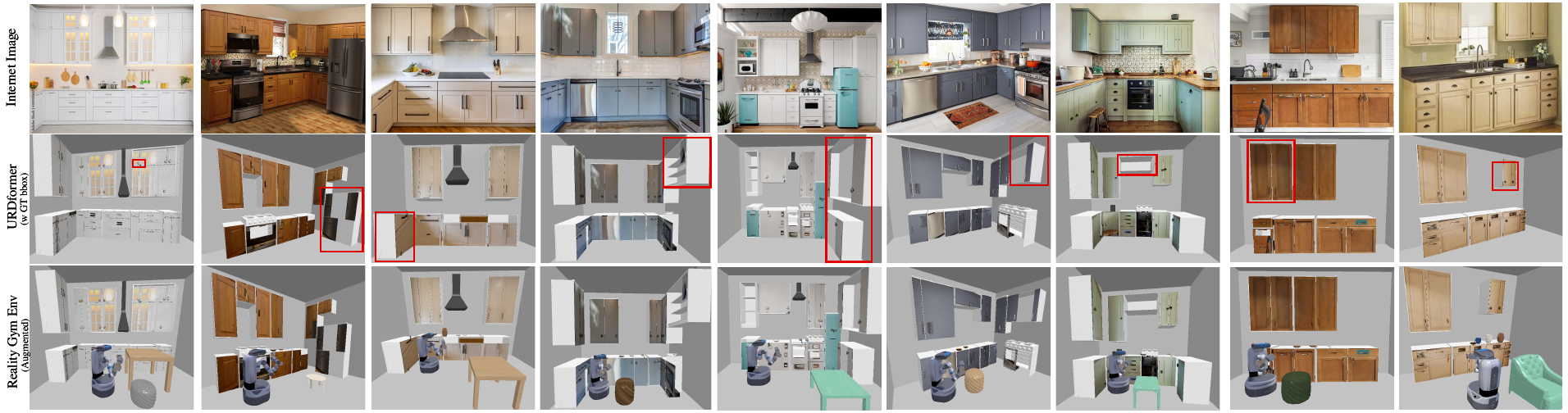}
    \caption{\textbf{Generated Kitchen Scenes:} Examples of kitchen scenes predicted by \method from internet images given labeled bounding boxes. Examples of failure parts are highlighted in red boxes. }
    \vspace{-1.5em}
    \label{fig:kitchen}
\end{figure*}

Overall, the proposed real2sim2real pipeline using \method and targeted randomization shows a \textbf{78\%} success rate across all cabinet variations and tasks and an \textbf{85\%} success rate on opening/closing tasks. Even though the \method's predictions are not always accurate, as indicated by the red boxes in Fig\ref{fig:real_world}, the targeted randomization robustifies the policy network to allow for zero-shot real-world transfer more effectively than un-targeted simulation generation.

\subsection{Can \method generate plausible and accurate simulation content from internet images?}



\subsubsection{Paired Training Dataset Generation}
\label{sec:datasetgenexp}

To synthesize our paired training dataset, we first procedurally generate a set of URDF representations of scenes in simulation both for global scenes like kitchens and for single objects like ovens, cabinets, and fridges (Fig~\ref{fig:data_generation}). 
We then follow the procedure in ~\ref{sec:paireddatasetgen} to control the data generation of paired images, generating a large dataset of simulation scenes and paired realistic RGB images (Fig~\ref{fig:data_generation}). For objects with diverse parts, we find that depth-guided Stable Diffusion~\citep{rombach2022high} often ignores the semantic details of local parts, leading to inconsistencies as visualized in the Appendix A-A). As described in Section~\ref{sec:paireddatasetgen} we generated a large and diverse set of texture templates by using images of existing textures downloaded from the internet to guide the depth-guided stable diffusion. During training, we then randomly chose one template texture and warped it back to the original part region using perspective transformation. Finally, we applied a stable diffusion in-painting model \cite{rombach2022high} to smooth the boundary of the parts and generate background content (Details are described in Appendix A-A). In total, we generated approximately 118K image-URDF pairs across 7 categories of single articulated objects, and approximately 200K image-URDF pairs of global kitchen scenes. 

\subsubsection{Object and Part Detection during Inverse Phase} \label{sec:Grounding_dino_appendix}
\method takes both the RGB image and bounding boxes of the object parts and predicts URDFs. During the inverse phase, we adopt an off-the-shelf open vocabulary object detector GroundingDINO \cite{liu2023grounding}.

\begin{figure}[h]
    \centering
    \includegraphics[width=0.48\textwidth]{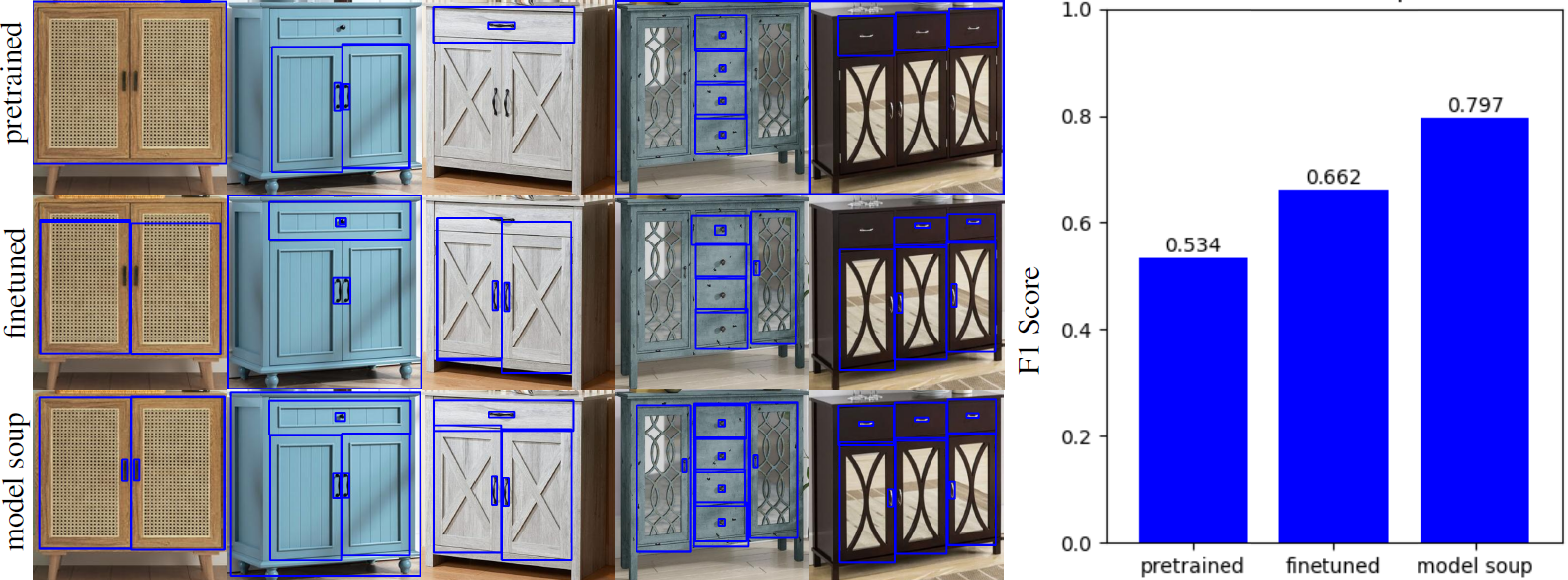}
    \caption{\footnotesize{Comparison among pretrained, finetuned and model soup GroundingDINO on cabinet dataset}}
    \vspace{-1.2em}
    \label{fig:model_soup}
\end{figure}
However, if we directly apply GroundingDINO on detecting parts such as drawers and handles, the detection performance is unsatisfying with an F1 score of  53.4\%. Instead, if we use the same generated dataset that was used to train URDFormer (Visualized in Appendix A) and finetune groundingDINO, we observe an improvement with an F1 score of 66.2\%. However, We also observe that compared to the pretrained GroundingDINO, the finetuned GroundingDINO often fails to detect parts that have unique shapes or patterns. Inspired by recent work Model Soup\cite{wortsman2022model}, we simply average the pretrained weights and the finetuned weights. This leads to surprising improvement, with an F1 score of 79.7\%. We additionally apply post-processing to remove duplicated boxes.  Figure \ref{fig:model_soup} shows Model Soup's influence on the bounding box detection.

\begin{figure*}[t]
    \centering
    \includegraphics[width=\textwidth]{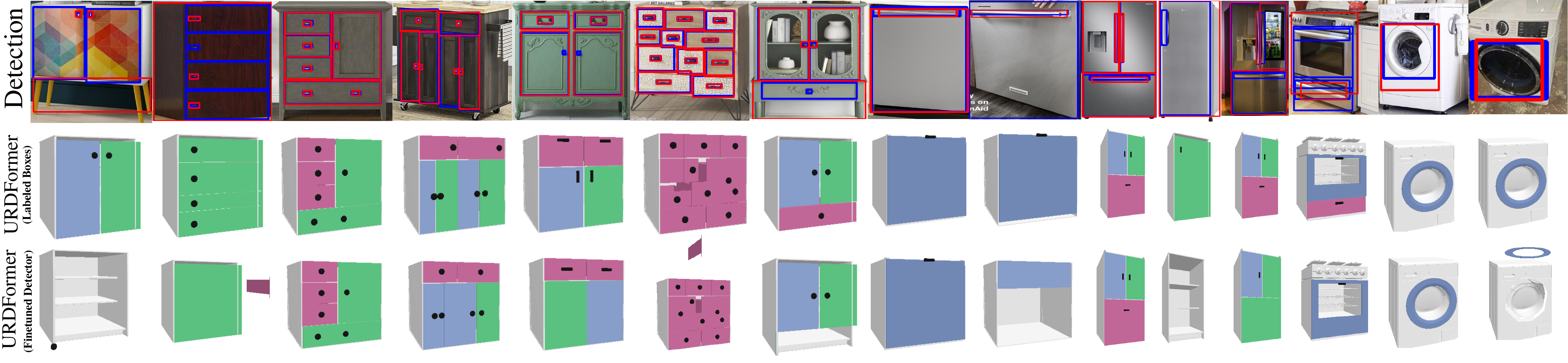}
    \caption{\footnotesize{Successful and unsuccessful examples comparing URDFormer prediction on different articulated objects using (1) manually labeled bounding boxes and (2) bounding boxes provided by fine-tuned GroundingDINO. Image results of fine-tuned GroundingDINO (\textcolor{red}{Red}) compared with manually labeled boxes (\textcolor{blue}{Blue}) are shown in the first row. Note that textures are removed for better visual comparison.}}
    \vspace{-0.2em}
    \label{fig:objects}
\end{figure*}
\begin{figure*}[t]
    \centering
    \includegraphics[width=\textwidth]{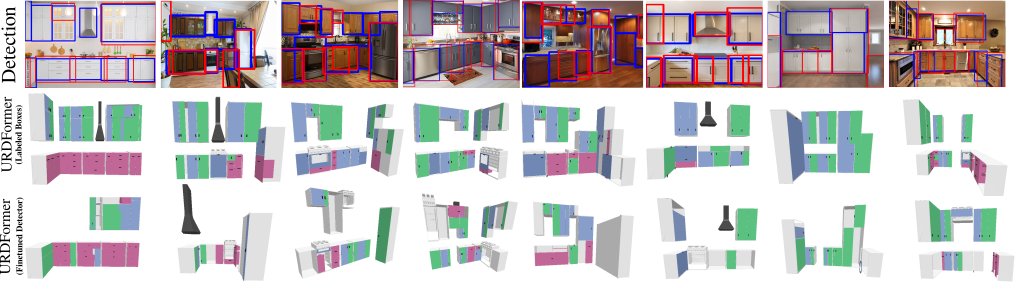}
    \caption{\footnotesize{Successful and unsuccessful examples on kitchens comparing URDFormer prediction based on boxes generated by (1) manual labeling and (2) fine-tuned GroundingDINO. The first row also shows the boxes detected by fine-tuned GroundingDINO (\textcolor{red}{Red}) compared with manually labeled boxes (\textcolor{blue}{Blue})}}
    \vspace{-1.3em}
    \label{fig:scenes}
\end{figure*}

\subsubsection{Real World Evaluation Datasets}

We create two types of test sets for the evaluation of \method: (a) \textit{Object-Level} set includes labeled URDFs of 300 internet images of individual objects from 5 categories including 100 cabinets, 50 ovens, 50 dishwashers, 50 fridges and 50 laundry machines. (b) \textit{Kitchen} set includes URDFs of 54 internet images of kitchens, with 5-15 articulated objects per kitchen. For each scene, we manually label the bounding box for each object and its parts, as well as the URDF primitives including mesh types, parent id, positions, and scales relative to its parent. We used the mesh types such as ``left door", and ``right door" to infer link axis and joint types. All the position values and scale values are discretized into 12 bins. 

\subsubsection{Evaluation Metrics} 

Evaluating entire scenes is challenging given the mixed structure and subjective nature of human labeling.  We therefore measured accuracy of the predicted model using three individual sub-tasks: category accuracy, parent accuracy and spatial error.  In order to compute these statistics, we must first align predicted objects in the scene to their ground truth counterparts.  This is accomplished by computing the intersection-over-union (IOU) of the 2D detected boxes used to instantiate the scene objects and the ground-truth bounding boxes of all the objects in the scene.  Hungarian matching is then used to assign each detected box to a single ground truth box.  Note that if the number of predicted boxes is not the same as the number of ground truth boxes, there will be some false positives (predicted objects that do not correspond to any ground truth objects) or false negatives (ground truth objects that do not have any associated predictions).  We therefore also record the overall precision and recall of the objects in the scene.  Once we have a set of aligned detected and ground-truth boxes, we say that the category of a predicted object is correct if it matches the category of the assigned ground truth object.  Similarly the predicted parent is correct if it matches the parent of the assigned ground truth object.  Finally the spatial error is the average absolute error of the predicted discretized spatial coordinates.  For larger objects, the model predicts four coordinates $x_1, y_1, x_2, y_2$ representing the bounding box of the object relative to its parent.  For small objects such as handles and knobs, we predict only the object center $x_1, y_1$ and so the spatial error only considers these two values.  Finally, we separate out these scores for high-level object predictions (cabinets, dishwashers, etc.) and low-level part predictions (doors, handles, etc.) so that the effects of various ablations on larger and smaller parts are more interpretable.

\subsubsection{Qualitative Results}
Figure \ref{fig:kitchen} shows several examples of kitchens generated by URDFormer using internet images.  While the model makes some mistakes, it is able to reproduce kitchen environments that largely match the structure of the original images.  Figures \ref{fig:objects} and \ref{fig:scenes} show successful and unsuccessful reconstructions of individual objects and scenes respectively.  These images also show results when ground-truth boxes are provided.  Again, while the model makes some mistakes, it largely captures the overall configuration of the objects and scenes.

\subsubsection{Ablation Study} 
\begin{table*}[]
\setlength{\tabcolsep}{3pt}
\begin{tabular}{cccccccccc|cccccccc}
 &
   &
  \multicolumn{8}{c|}{Kitchen} &
  \multicolumn{8}{c}{Object-Level} \\ \cline{2-18} 
 &
   &
  \multicolumn{4}{c|}{GT boxes} &
  \multicolumn{4}{c|}{Finetuned Grounding DINO} &
  \multicolumn{4}{c|}{GT Boxes} &
  \multicolumn{4}{c}{Finetuned Grounding DINO} \\ \cline{2-18} 
 &
   &
  Ours &
  Random &
  Sim &
  \multicolumn{1}{c|}{Selected} &
  Ours &
  Random &
  Sim &
  Selected &
  Ours &
  Random &
  Sim &
  \multicolumn{1}{c|}{Selected} &
  Ours &
  Random &
  Sim &
  Selected \\
\multirow{5}{*}{\rotatebox[origin=c]{90}{Global}} &
  Mesh Acc ($\uparrow$) &
  \textbf{0.578} &
  0.42 &
  0.407 &
  \multicolumn{1}{c|}{0.576} &
  \textbf{0.603} &
  0.533 &
  0.354 &
  0.462 &
  \textbf{0.740} &
  0.463 &
  0.520 &
  \multicolumn{1}{c|}{0.677} &
  \textbf{0.688} &
  0.440 &
  0.520 &
  0.630 \\
 &
  Parent Acc ($\uparrow$) &
  \textbf{0.833} &
  0.831 &
  0.813 &
  \multicolumn{1}{c|}{0.807} &
  0.816 &
  \textbf{0.819} &
  0.810 &
  \textbf{0.819}&
  --- &
  --- &
  --- &
  \multicolumn{1}{c|}{---} &
  --- &
  --- &
  --- &
  --- \\
 &
  Spatial Err ($\downarrow$) &
  \textbf{0.809} &
  0.845 &
  0.848 &
  \multicolumn{1}{c|}{0.885} &
  \textbf{0.987} &
  1.036 &
  1.038 &
  1.032 &
  --- &
  --- &
  --- &
  \multicolumn{1}{c|}{---} &
  --- &
  --- &
  --- &
  --- \\
 &
  Recall ($\uparrow$)&
  1 &
  1 &
  1 &
  \multicolumn{1}{c|}{1} &
  0.726 &
  0.726 &
  0.726 &
  0.726 &
  --- &
  --- &
  --- &
  \multicolumn{1}{c|}{---} &
  --- &
  --- &
  --- &
  --- \\
 &
  Precision ($\uparrow$)&
  1 &
  1 &
  1 &
  \multicolumn{1}{c|}{1} &
  0.951 &
  0.951 &
  0.951 &
  0.951 &
  --- &
  --- &
  --- &
  \multicolumn{1}{c|}{---} &
  --- &
  --- &
  --- &
  --- \\ \hline
\multirow{5}{*}{\rotatebox[origin=c]{90}{Parts}} &
  Mesh Acc ($\uparrow$)&
  0.704 &
  \textbf{0.719} &
  0.662 &
  \multicolumn{1}{c|}{0.675} &
  0.537 &
  \textbf{0.558} &
  0.505 &
  0.522 &
  0.903 &
  \textbf{0.927} &
  0.867 &
  \multicolumn{1}{c|}{0.861} &
  0.851 &
  \textbf{0.865} &
  0.811 &
  0.803 \\
 &
  Parent Acc ($\uparrow$)&
  0.765 &
  \textbf{0.773} &
  0.763 &
  \multicolumn{1}{c|}{0.75} &
  \textbf{0.711} &
  0.696 &
  0.707 &
  0.711 &
  0.874 &
  \textbf{0.878} &
  0.857 &
  \multicolumn{1}{c|}{0.857} &
  \textbf{0.826} &
  0.821 &
  0.806 &
  0.805 \\
 &
  Spatial Err ($\downarrow$)&
  1.799 &
  \textbf{1.699} &
  1.891 &
  \multicolumn{1}{c|}{1.981} &
  2.957 &
  2.885 &
  \textbf{2.798} &
  3.107 &
  0.478 &
  \textbf{0.420} &
  0.649 &
  \multicolumn{1}{c|}{0.867} &
  0.791 &
  \textbf{0.753} &
  0.917 &
  1.129 \\
 &
  Recall ($\uparrow$)&
  1 &
  1 &
  1 &
  \multicolumn{1}{c|}{1} &
  0.495 &
  0.495 &
  0.495 &
  0.495 &
  1 &
  1 &
  1 &
  \multicolumn{1}{c|}{1} &
  0.853 &
  0.832 &
  0.832 &
  0.832 \\
 &
  Pred Precision ($\uparrow$)&
  1 &
  1 &
  1 &
  \multicolumn{1}{c|}{1} &
  0.927 &
  0.927 &
  0.927 &
  0.927 &
  1 &
  1 &
  1 &
  \multicolumn{1}{c|}{1} &
  0.986 &
  0.987 &
  0.987 &
  0.987

\end{tabular}
\caption{\textbf{Ablation Study}. We analyze which part training with generated texture benefits the most by comparing URDFormer trained with other textures, across GT boxes and boxes from the finetuned Grounding DINO detector.}
\vspace{-1.5em}
\label{tab:ablation_table}
\end{table*}

\begin{figure*}[t]
    \centering
    \includegraphics[width=0.9\textwidth, height=0.16\textheight]{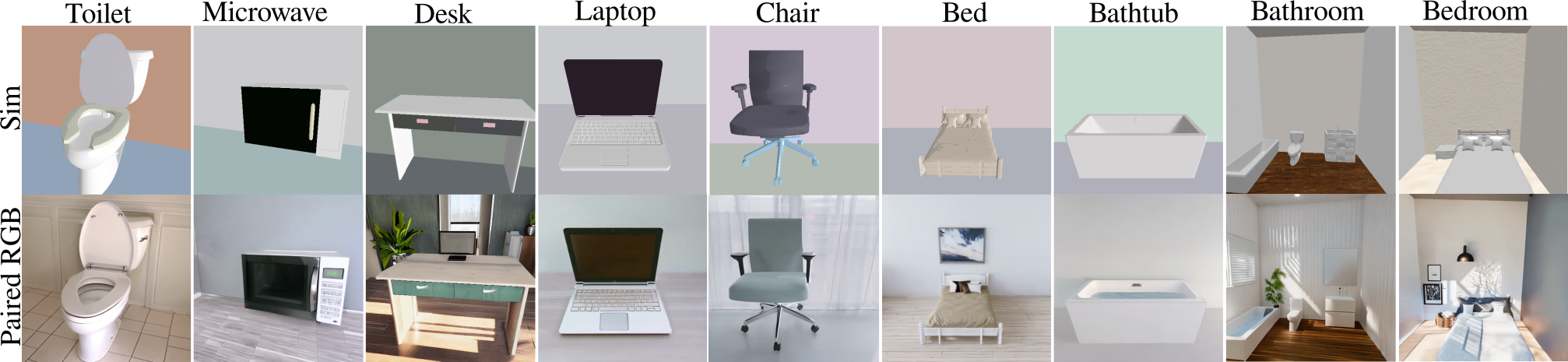}
    \vspace{-0.5em}
    \caption{The forward phase can be applied to additional objects and create a diverse dataset for training \method. }
    \vspace{-0.5em}
    \label{fig:new_dataset}
\end{figure*}

\begin{figure*}[t]
    \centering
    \includegraphics[width=\textwidth, height=0.33\textheight]{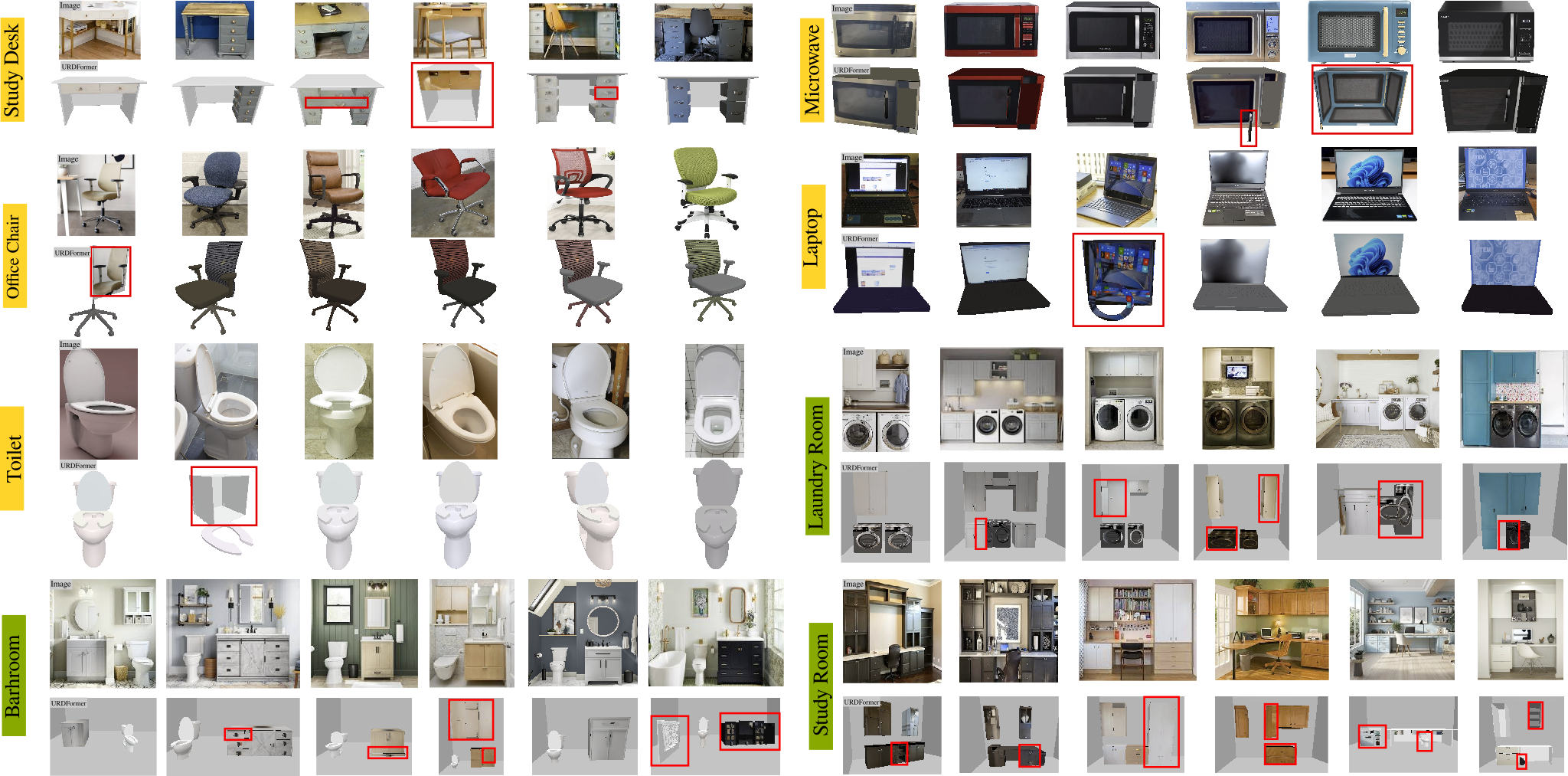}
    \vspace{-1em}
\caption{\footnotesize{Qualitative study of \method generalization to other scenes and objects with successful and unsuccessful (highlighted in red) examples. In particular, We train \method on the new dataset with added categories shown in Fig\ref{fig:new_dataset} and evaluate on internet images for both object-level and global prediction. Interestingly, we did not train \method on the Laundry Room and Study Room but found \method can generalize to these unseen categories, which is likely due to the two-stage training of \method Global and \method Part. Please note that \method does not reconstruct accurate meshes or predict properties such as friction or mass, which is detailed in the limitation and future work sections.}}
    \vspace{-1.5em}
    \label{fig:new_prediction}
\end{figure*}

We compare the prediction of URDFormer trained with (1) \textbf{Ours}: realistic texture generated by part-consistency stable diffusion (2) \textbf{Random}: random texture downloaded from Describable Textures Dataset \cite{cimpoi14describing} (3) \textbf{Sim}: random 3-channels RGB color and (4) \textbf{Selected}: carefully selected texture images that matches the object categories, such as wood texture (cabinet), metal texture (dishwasher) and so on. The results are shown in Table \ref{tab:ablation_table}. We observe that generated realistic texture is particularly helpful in global prediction, including identifying object types (cabinet or oven), parents (which wall the cabinet belongs to), and where to put the object. Surprisingly, we found the texture realism does not affect so much when predicting part structures and sometimes is slightly worse than random texture by 1\% to 2\%. This is likely due to using bounding box position features is sufficient for predicting simple low-level structures. For example, if a small box A is in the center of another box B, box B is likely a drawer instead of a door or a handle. On average, prediction using finetuned object detector performance is worse than using GT boxes due to detection error. However, one surprising observation is that using detected boxes helps slightly with identifying global object types. We hypothesize this is because boxes labeled by humans sometimes group multiple cabinets into one single box if they are close, making mesh prediction slightly challenging.

\subsubsection{Reality Gym}
With the ability to cheaply generate an arbitrary number of realistic simulation assets directly from the internet, we introduce \textit{RealityGym}, a robot learning suite with a collection of realistic simulation assets and scenes created from real world RGB images using \method. We provide an initial set of 300 objects (Cabinets, Ovens, Fridges, wahsers and Dishwashers) and 50 kitchen scenes from internet images, with future work looking to expand this into a bigger dataset. In addition, we also provide 84 meshes of cabinet frames, 20 door meshes, 59 drawers, 440 handles and 116 knobs from PartNet\cite{mopartnet}. We can randomly incorporate these meshes into the initial URDFs to generate diverse scenes complete with articulated objects. We define 4 main tasks: (1) Open any articulated parts \ie top middle drawer (2) Close any articulated parts (3) Fetch objects (4) Collect objects. We automatically generate tasks and their language descriptions, and use a motion planner (Curobo~\cite{curobo_report23}) to complete the tasks. Fig~\ref{fig:reality_gym_suite} and Fig~\ref{fig:kitchen} show examples of robots performing in RealityGym on a variety of generated simulation environments and assets. Details about RealityGym can be found in Appendix C.  

\subsection{Can \method generalize to diverse objects and scenes?}
\label{sec:rebuttal_generalization}

In order to demonstrate the generalization capability of \method, we used the same techniques discussed on Section \ref{sec:method} to create five additional object categories and four additional scene categories for qualitative evaluation.  The new object categories are toilet, microwave, desk, laptop and chair, while the four scene categories are bedroom, bathroom, laundry room and study room.  We trained a single new part model that incorporates these additional categories by adding approximately 6k training examples per object category to the original training dataset.  We took the same approach to train a single global scene model and added approximately 10k training examples for the bathroom and bedroom.  We found that the laundry room and study could be adequately captured with no new global scene examples, as these categories only contain objects that are present in the other scene categories.  Figure \ref{fig:new_dataset} shows example training data generated by the forward pipeline, while Figure \ref{fig:new_prediction} shows qualitative examples of objects and scenes inferred by a \method trained on this data.  More examples are available in Appendix F.

\subsection{Can \method support different robots and tasks?}

\label{sec:rebuttal_robot}

\begin{figure}[!h]
  \centering
  \includegraphics[width=0.45\textwidth]{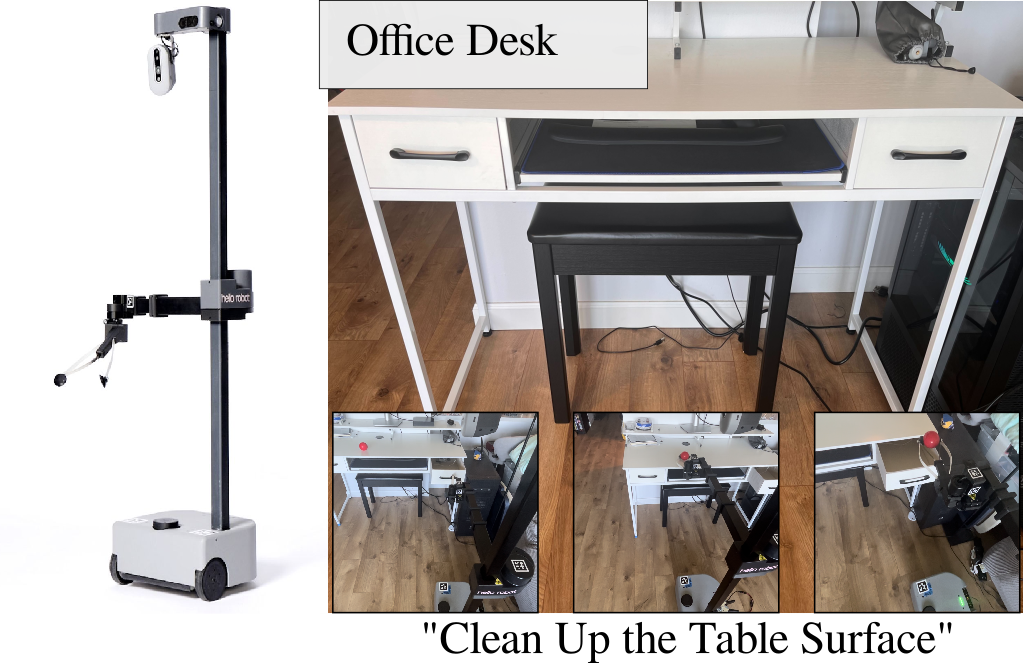}
  \caption{We trained a Stretch robot on a multi-step task "Clean Up the Table Surface" using \method prediction}
  \vspace{-1.5cm}
  \label{fig:new_robot_a}
\end{figure}

\begin{figure*}[t]
    \centering
    \includegraphics[width=\textwidth]{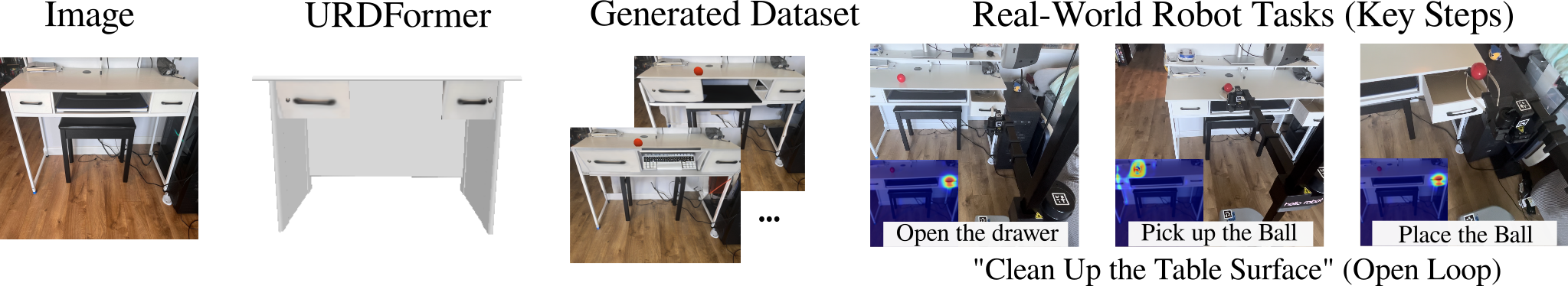}
    \vspace{-1.5em}
\caption{\footnotesize{\method} can be applied to a different robot such as a Stretch Robot to perform a multi-step task such as "clean up the table surface". We apply URDFormer to predict the URDF of a study desk, and render a dataset to train a vision-based policy to predict affordance map at each step.}
    \vspace{-2em}
    \label{fig:new_robot}
\end{figure*}

To demonstrate that \method supports multi-step tasks and different robots, we train an additional policy for a Stretch robot to place an object in a desk drawer as shown in Figure \ref{fig:new_robot_a}. Figure \ref{fig:new_robot} shows the data generation process for training the mobile robot's policy.  First, a single image of the robot's environment is passed to a pretrained URDFormer model to produce a scene description of the desk.  Then, this scene description is used to generate training data in simulation that can instruct the robot how to accomplish it's multi-stage objective.  When generating this data, we use an inpainting model \cite{rombach2022high} to reduce the sim2real gap by inpainting the pixels covered by the simulated object overlaid onto the original image.  Finally the robot policy, which predicts a per-step affordance map from an initial image, is trained on this new generated data.  This policy is implemented as a UNet, which takes the initial observation, as well as the task embedding and predicts multiple object affordance maps.  These affordance maps are used with a motion planner to guide the robot at each step.  Details on training the policy can be found in Appendix D-C.

\section{Related Work}
\label{sec:related_work}

\subsection{Asset Creation for Robot Manipulation}

Constructing realistic and diverse assets for robot learning and control has been an important goal for many years.  Early efforts \cite{izadi2011kinectfusion, agarwal2011building, henry2012rgb, mur2015orb} used geometric approaches to reconstruct static 3D scenes.  Recently learned approaches \cite{park2019deepsdf, mildenhall2021nerf, kerbl20233d} have in some cases improved the accuracy and visual realism of these approaches.  While these techniques can offer incredible visual realism, they do not usually produce articulated models that allow for dynamic interaction.

%
To address this, many
prior works have explored utilizing assets that are manually curated~\cite{kolve2017ai2,xiang2020sapien,szot2021habitat,deitke2023objaverse}, procedural-generated~\cite{deitke2022,yang2023holodeck}, or scanned from the real world~\cite{martin2019rbo,xia2020interactive,liu2022akb}, to train robot manipulation policies.
While the learned policies are supposed to apply directly to test scenes, there could be challenging scenes in which the systems fail to faithfully recognize and interact with the scene due to the scene complexity or sim-to-real gaps.
In our work, we resort to a targeted system that first reconstructs a URDF given the test scene and then performs scene-specific training to learn a specialized model in the scene.
There is also a rich body of literature on building digital twins of articulated objects from the real world, either based on passive visual observations such as 2D multi-view images or videos~\cite{yang2021lasr,Qian22,wei2022self,heppert2023carto,liu2023paris}, 3D RGB-D, depth images or point clouds~\cite{wang2019shape2motion,yan2020rpm,weng2021captra,abdul2022learning, heiden2022inferring}, full-scene 3D scans~\cite{li21openrooms, mao2022multiscan, deitke23phone2proc}, or through interactive perception which requires few-shot interactions with the object~\cite{katz2013interactive, urgenlearning, nie2022structure, jiang2022ditto, ma2023sim2real, hsu2023ditto, memmel2023asid}.
While in some cases, these methods can successfully produce high quality models, they require either video, depth sensing or interaction with the objection which is more difficult to aquire and is rarely available in online internet data.
Our work, instead, advocates for a faster and cheaper pipeline where we reconstruct a simple URDF model, which is sufficient for downstream learning, from a single image without the need for laborious scanning or object interaction.

\subsection{Synthetic Data Augmentation with Generative Models}

Everyone can freely generate realistic image content nowadays with the booming development of the latest generative AI technology~\cite{ho2020denoising,rombach2022high}.
Researchers have also explored the use of such powerful generative models in creating large-scale high-quality synthetic data for training perception systems in different applications, such as image classification~\cite{azizi2023synthetic,trabucco2023effective}, object segmentation~\cite{li2023open,yu2023diffusion,wu2023datasetdm}, and representation learning~\cite{fu23dreamsim, tian23stablerep, jahanian22genmodels}.
In the field of robotics, several studies have attempted to reduce sim-to-real gaps and broaden the domain coverage using synthetic data augmented with generative models~\cite{ho2021retinagan,chen23genaug, mandi2022cacti, yu23rosie, bharadhwaj2023roboagent, trabucco23dataaug}.
In our case, we look to augment synthetic asset renderings with realistic textures produced by generative models.
While recent studies have proposed controllable image generation algorithms such as ControlNet~\cite{zhang2023adding} and 3D mesh texturing methods~\cite{richardson2023texture,chen2023scenetex}, 
we find them lacking in preserving part-level structure details and consequently propose a novel controlled texture generation method to produce realistic synthetic data for training.

\section{Limitations and Future Work}
\label{sec:limitations}
Please see the full list of limitations and future work in Appendix F.

\textbf{Part Detection}
\method relies on the performance of bounding box detection. Although the finetuned Grounding DINO improves performance than the pretrained model, there is still a gap for improvement, especially on global scene detection.

\textbf{Texture and Meshes} \method focuses on predicting kinematic URDF structures and uses predefined meshes that might not match the real-world scenes. To apply textures, we simply assume all parts are rectangular shapes, and use the bounding box of each object part to crop the image and import it into a uv map template. However, this does not work for irregular meshes such as a donut-shape door, or when the object in the image is tilted.

\textbf{Limited URDF Primitives}
\method currently only supports articulated objects that have limited joint types such as prismatic and revolute, and cannot predict complex objects such as cars and lamps. 

\textbf{Link Collisions} \method only predicts URDF primitives for each bounding box, which sometimes leads to a collision between two links. Further post-processing is required to resolve this issue. 

\textbf{Multiple Trained Components} Our pipeline is not trained end-to-end and consists of multiple learning components.  While this increases the complexity of the system, it is necessary to ensure consistency when using the generative model, and make the most use of existing pretrained components.

\textbf{Inferred Physical Properties} Our system does not presently attempt to infer physical properties such as mass, inertial moments or friction directly from observations in the scene.  In theory the visual information present in the images may allow for a rough approximations of these quantities.  We expect this to be a productive direction for future work.
\section{Conclusion}
We present a scalable pipeline for creating articulated simulation assets from real-world images. In particular, we introduce a forward-inverse framework that generates realistic and consistent images of articulated objects and trains a transformer-based network URDFormer to predict their corresponding URDFs. We present RealityGym, a robot learning suite with realistic simulation assets generated from real-world images. Additionally, we show that the predicted URDFs with targeted domain randomization enable better zero-shot performance in the real-world. Our pipeline provides the first step towards cheaper and scalable realistic scene generation for robot learning. 
\section{Acknowledgement}
We thank all members of UW Robotics Lab for all the support and discussions. In particular, we thank all the lovely people from WEIRD Lab and RSE lab for insightful feedback. We especially thank Mateo Guaman Castro for his detailed feedback on the paper draft, and Marcel Torne
for fruitful discussion. We also thank Microsoft for UR5 robots, and Adam Fishman, Rohan Baijal, Tyler Han and Yuquan Deng for helping me move these two heavy robots, and Jacob Berg for letting me borrow his real-sense camera cable. The work was also funded in part by Amazon Science Hub. 

\clearpage

\appendices

\section{Paired Dataset Generation for Training \method}
\subsection{Consistency-aware Image Generation}
\begin{figure}[!h]
    \includegraphics[width=0.49\textwidth]{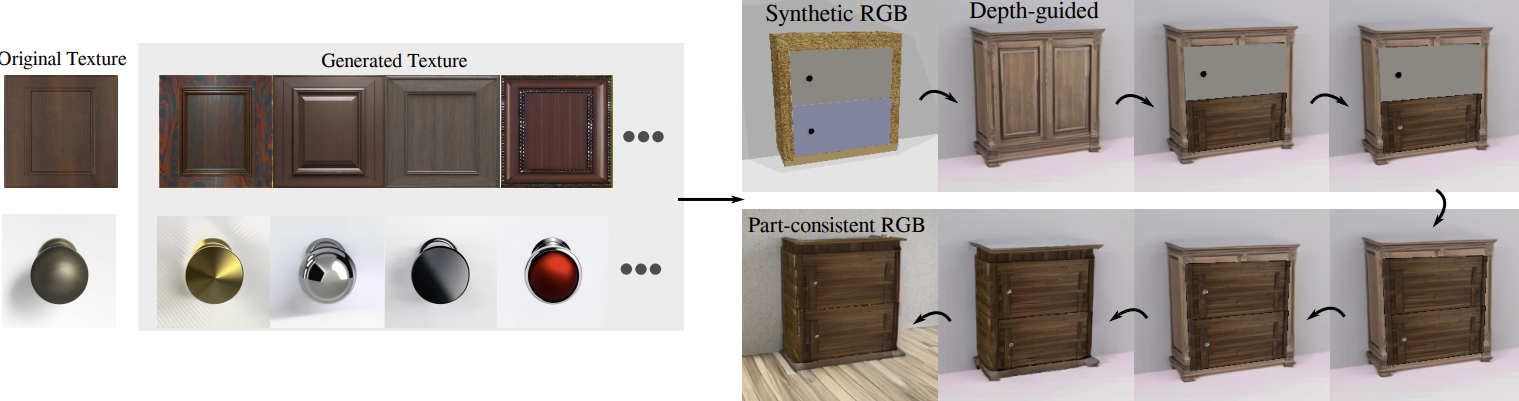}
\caption{\footnotesize{Paired dataset generation using texture and prompt templates to guide Stable Diffusion \citep{rombach2022high} and create a diverse texture dataset, which can be then warped on the targeted individual part of the object.}}
\label{fig:part_render}
\end{figure}

\begin{figure}[!h]
    \includegraphics[width=0.49\textwidth]{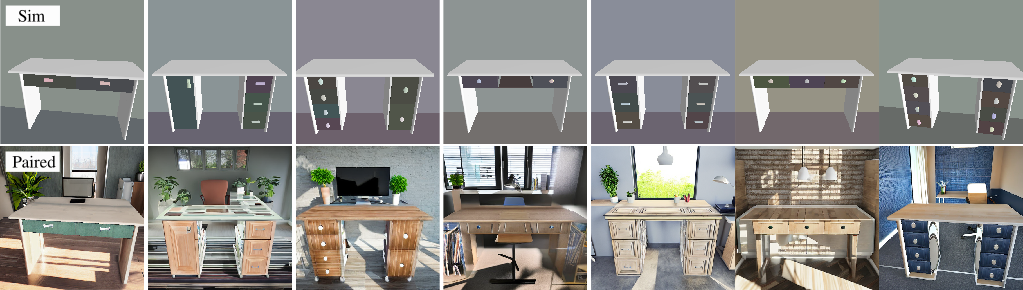}
\caption{\footnotesize{Our forward phase can generate diverse configurations while maintaining local consistency.}}
\label{fig:diverse_config}
\end{figure}

We observe that depth-guided or in-painting stable diffusion models \cite{rombach2022high} often ignore local consistency, making it difficult to render high-quality images that are paired with the simulation content. To overcome this issue, we propose texture-guided image generation. Instead of asking stable diffusion to texture the entire object which might change the low-level details, we utilize the advantage of the diffusion models that help to diversify an initial small set of texture images. In particular, given an initial 100 cabinet texture images we downloaded from the internet, we apply stable diffusion to change the style, pattern, or color, which significantly increases the texture dataset. Then we randomly choose a texture and simply apply it to regions of interest (\ie drawer, door) obtained by the simulator using perspective warping. One important note is that these texture images are not UV texture maps, instead they are simply 2D images, and the texturing step is only at 2D image space because we only care about the photo realism for that particular image. As shown in Fig \ref{fig:part_render}, this simple approach is surprisingly effective and creates realistic images while maintaining low-level part consistency. We can randomly generate a dataset with diverse configurations as shown in Figure \ref{fig:diverse_config}

\subsection{Part-Consistency}
\label{sec:part-consistency_appendix}
\begin{figure}[!h]
    \centering
    \includegraphics[width=0.49\textwidth]{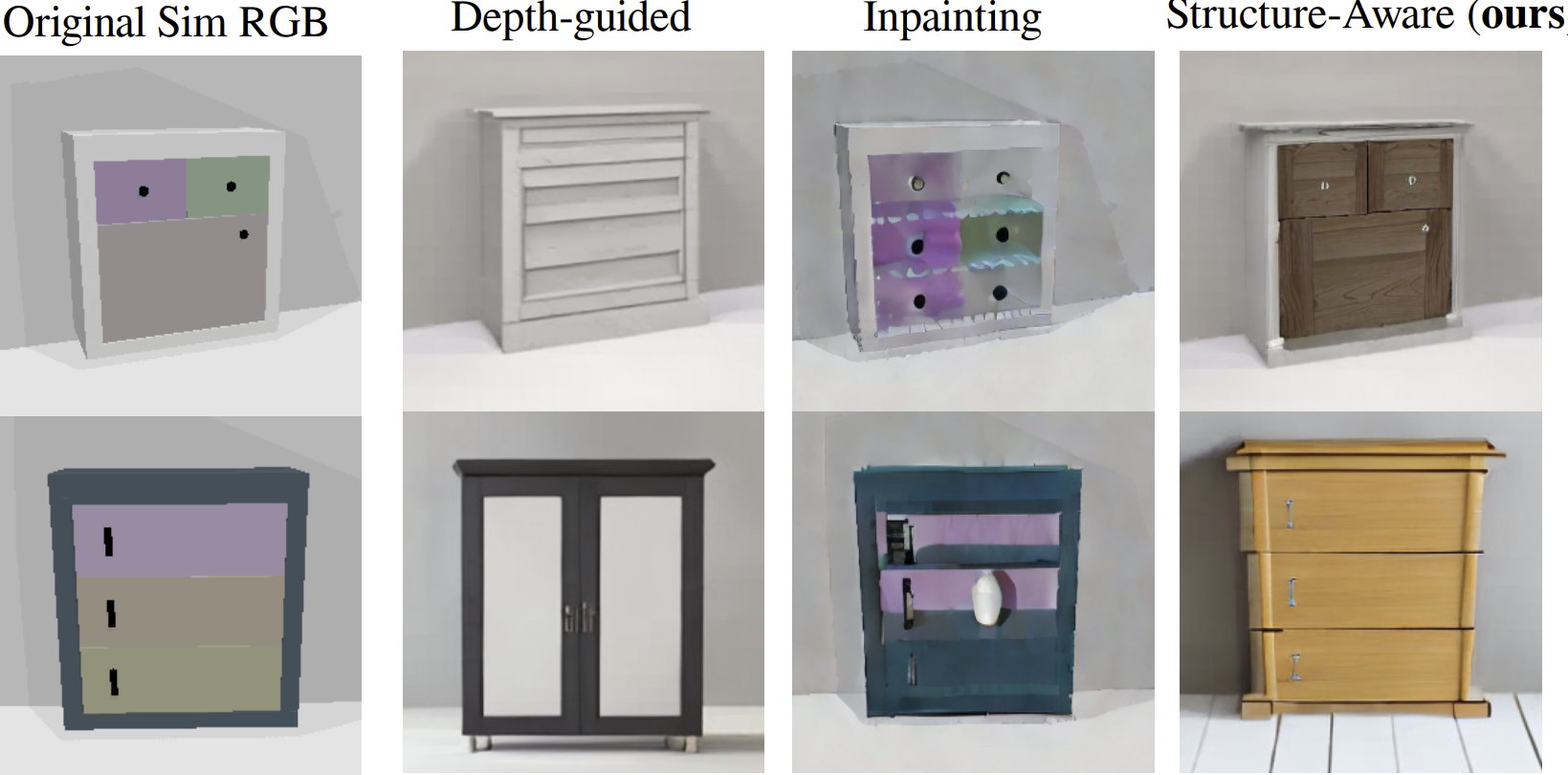}
    \caption{Qualitative comparison among different rendering methods: depth-guided diffusion models, inpainting stable diffusion and part-wise generation}
    \label{fig:render_baseline}
\end{figure}

We compare our part-wise generation method with other approaches qualitatively in Fig \ref{fig:render_baseline}. off-the-shelf stable diffusion models often ignore low-level details, making it challenging to create high-quality accurately paired dataset. Instead, our approach helps to preserve semantic low-level details while creating photorealistic paired images that match the original textureless simulation data. 


\begin{figure}[!h]
    \centering
    \includegraphics[width=0.49\textwidth]{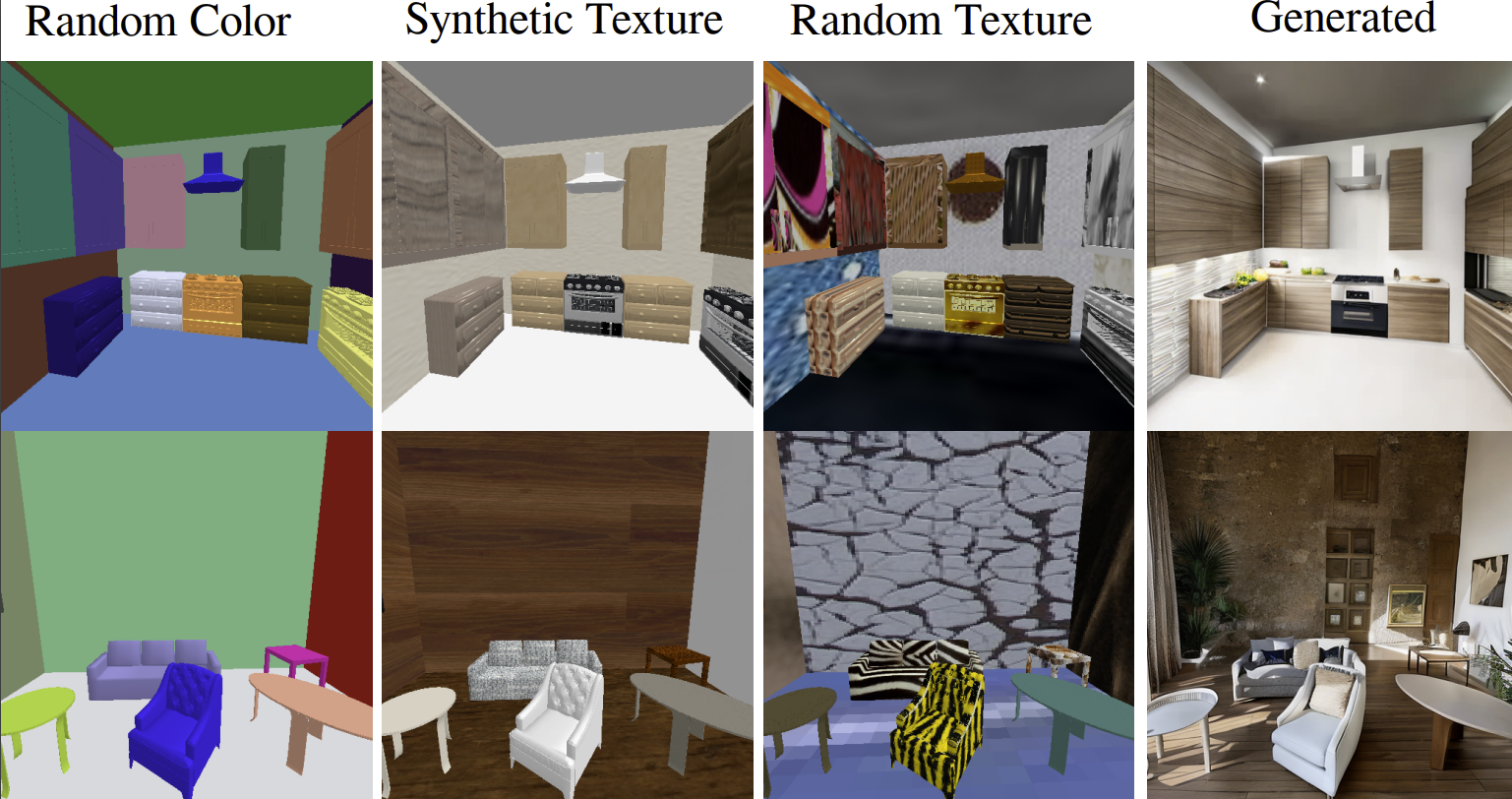}
    \caption{Visual comparison for ablation study with different training input: Random Colors, selected textures, random textures and generated textures. Generated textures show photo-realism that closer to the real-world distribution.}
    \label{fig:input_baselines}
\end{figure}

\begin{figure*}[t]
    \centering
    \includegraphics[width=\textwidth]{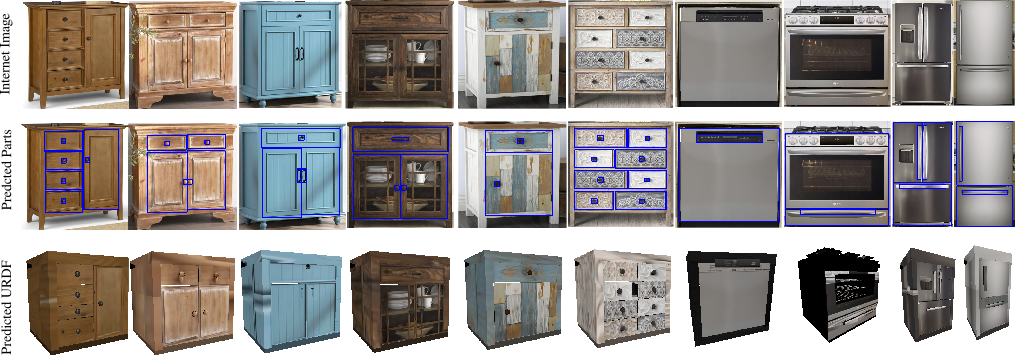}
    \caption{Reality Gym Assets: Converting real-world images into fully interactive simulation assets. Given internet images, we apply finetuned groundingDINO to obtain 2D part boxes. These boxes and RGB images are fed into URDFormer to generate articulated simulation assets that match the real-world structure.
    }
    \label{fig:reality_gym_appendix}
\end{figure*}
\label{sec:part-consistent_appendix}
\subsection{Visualization on Ablations}
\label{sec:part-consistent_qualatitive_appendix}
We visualize the different training data shown in Table II: \method with random colors, selected textures, random textures, and generated textures. All training inputs are captured from the same camera angles. As shown in Fig \ref{fig:input_baselines}, the generated texture shows high pixel realism that is closer to the distribution of the real world. As shown in Table II, training on such data improves performance in predicting mesh classes such as identifying ovens or cabinets.

\section{\method}
\label{sec:urdformer_appendix}
\subsection{Dataset}
\label{sec:dataset_appendix}
In total,
we generated approximately 118K image-URDF pairs across
5 common categories of articulated objects in the kitchen including cabinets, ovens, dishwashers, washer, fridges, as well as 2 types of rigid objects oven fans and shelves. These articulated objects include part meshes in 8 types: drawer, left door, right door, oven door, down door (dishwasher), circle door, handle and knob. 

\subsection{Training Details}
\label{sec:baseline_urdformer_appendix}
\method  is trained on one A40 GPU with batch-size of 256. For the ablation study shown in Table II, all the methods are trained with an equal number of epochs and evaluated using the last checkpoint. In particular, we train 130 epochs for global scenes and 80 epochs for objects. 



\begin{figure}[htb]
    \centering
    \includegraphics[width=0.45\textwidth]{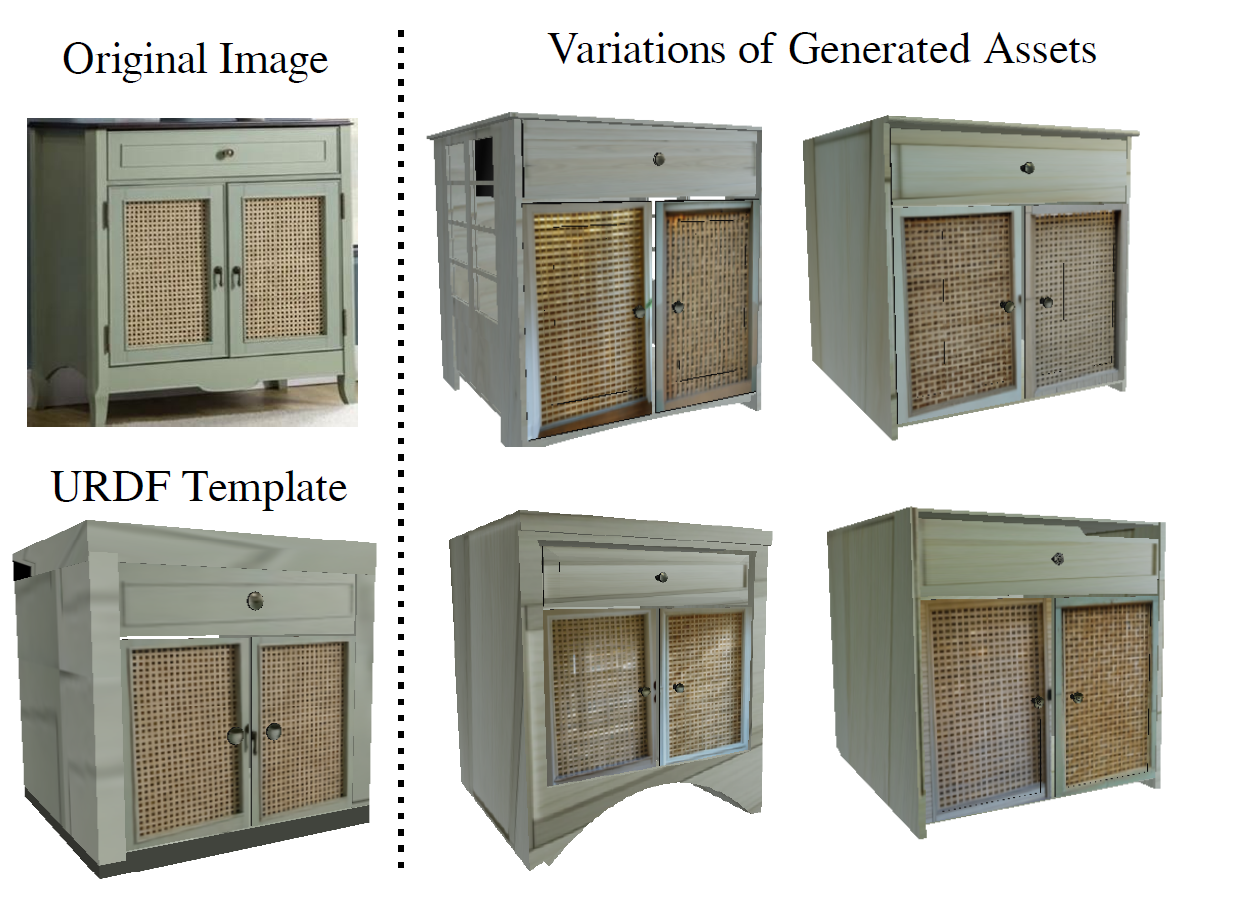}
    \caption{URDFormer does not reconstruct meshes. To cover the distribution of the real world, we randomly choose meshes from the PartNet dataset to replace the original part meshes for handles, drawers and so on.}
    \label{fig:mesh_randomization}
\end{figure}

\begin{figure}[htb]
    \centering
    \includegraphics[width=0.45\textwidth]{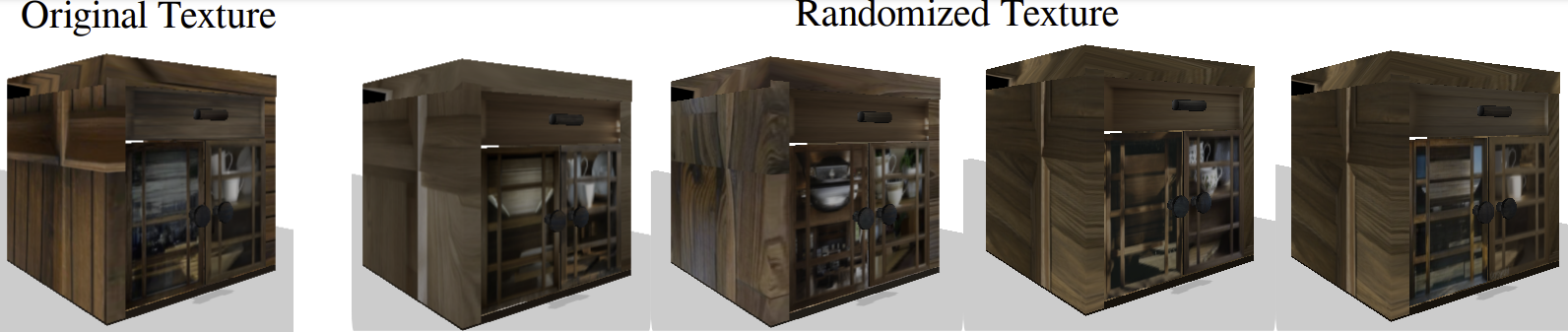}
    \caption{We randomize the texture of the object using stable diffusion to cover the distribution of the real world.}
    \label{fig:texture_randomization}
\end{figure}

\section{Reality Gym}
\label{sec:reality_gym_appendix}

\subsection{Realistic Assets from Images}
\label{sec:assets_appendix}
Fig \ref{fig:reality_gym_appendix} shows examples of assets predicted from internet images. Given an RGB image, we first apply finetuned GroundingDINO (model soup applied) to obtain 2D bounding boxes for each part. These boxes and the original RGB image are fed into URDFormer to predict the corresponding URDF which matches the kinematic structure of the image. We introduce 4 main tasks (1) Open any part (2) Close any part (3) Fetch the object (4) Collect the object, and automatically generate successful demonstrations using Curobo \cite{curobo_report23} and their corresponding language descriptions.  

\subsection{Targeted Domain Randomization}
\label{sec:tr_appendix}
To covert the real-world distribution, we further apply targeted domain randomization. This includes (1) Mesh Randomization: we randomly replace the original meshes with meshes from the PartNet \cite{mopartnet} dataset for parts such as doors, handles as well as cabinet frames, shown as Fig \ref{fig:mesh_randomization}(2) Texture Randomization: we use stable diffusion to generate different style of cropped texture images to form UV texture maps, shown as Fig \ref{fig:texture_randomization}. 

\begin{figure}[htb]
    \centering
    \includegraphics[width=0.4\textwidth]{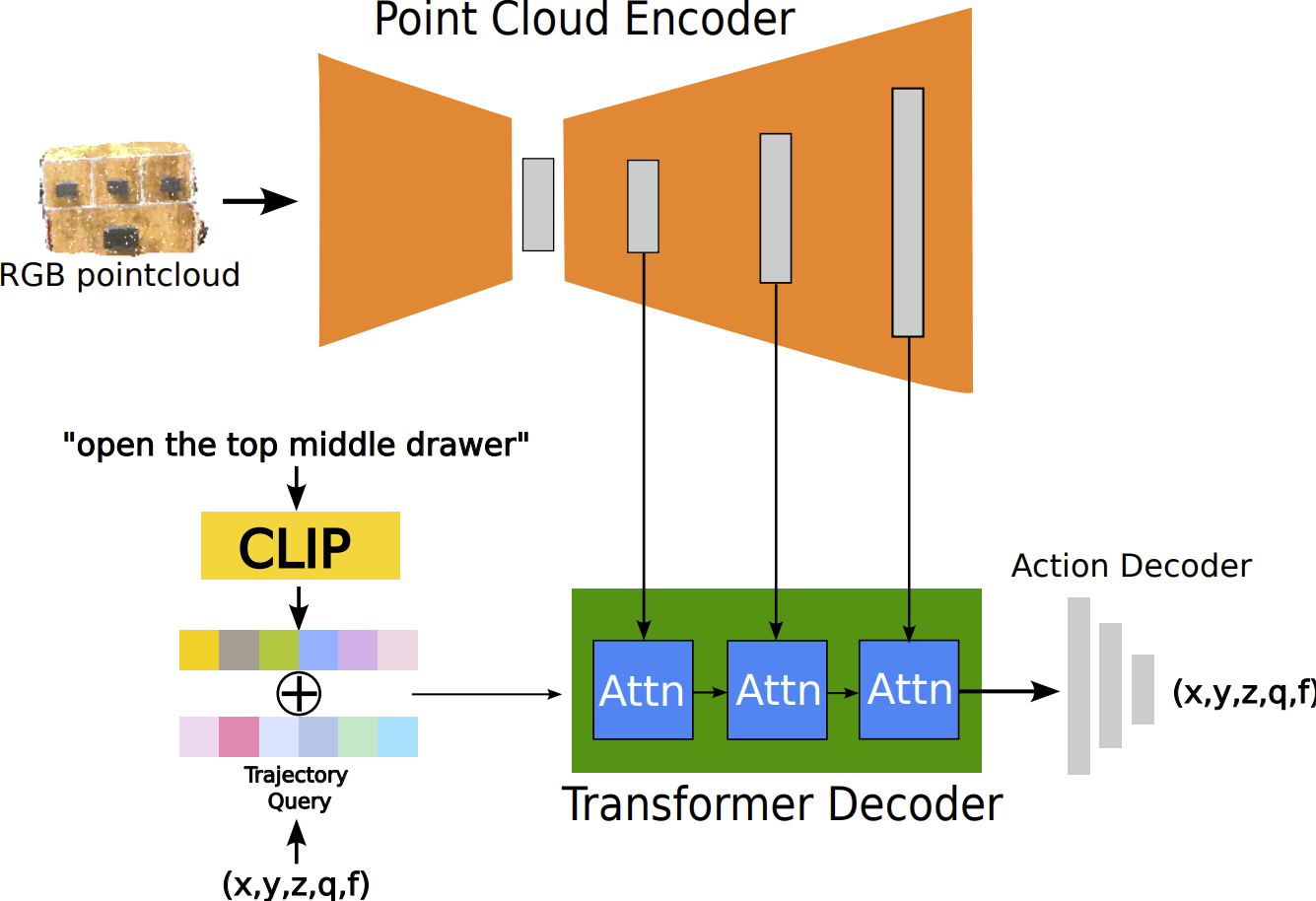}
    \caption{\footnotesize{We train language-conditioned policy based on M2T2 network structure. The network takes RGB pointcloud and predicts endeffector poses to complete the task.}}
    \label{fig:bc_network}
\end{figure}

\section{Real-World Experiments}
\label{sec:real_world_appendix}

\subsection{Data Collection in Sim}
\label{sec:sim_tasks_appendix}
We introduce 4 different tasks and distribute 2 tasks for each object. (1) Open any articulated part (2) Close any articulated part (3) Fetch the object from a particular drawer (4) collect the object into a particular drawer. For each task, we collect 500 demonstrations in simulation with objects in different poses. During training, we additionally apply online pose augmentation, as well as color augmentation for RGB pointcloud to reduce the sim2real gap. 

\subsection{Visualization of OWL-VIT}
\label{sec:owl_vit}
\begin{figure}[htb]
    \centering
    \includegraphics[width=0.45\textwidth]{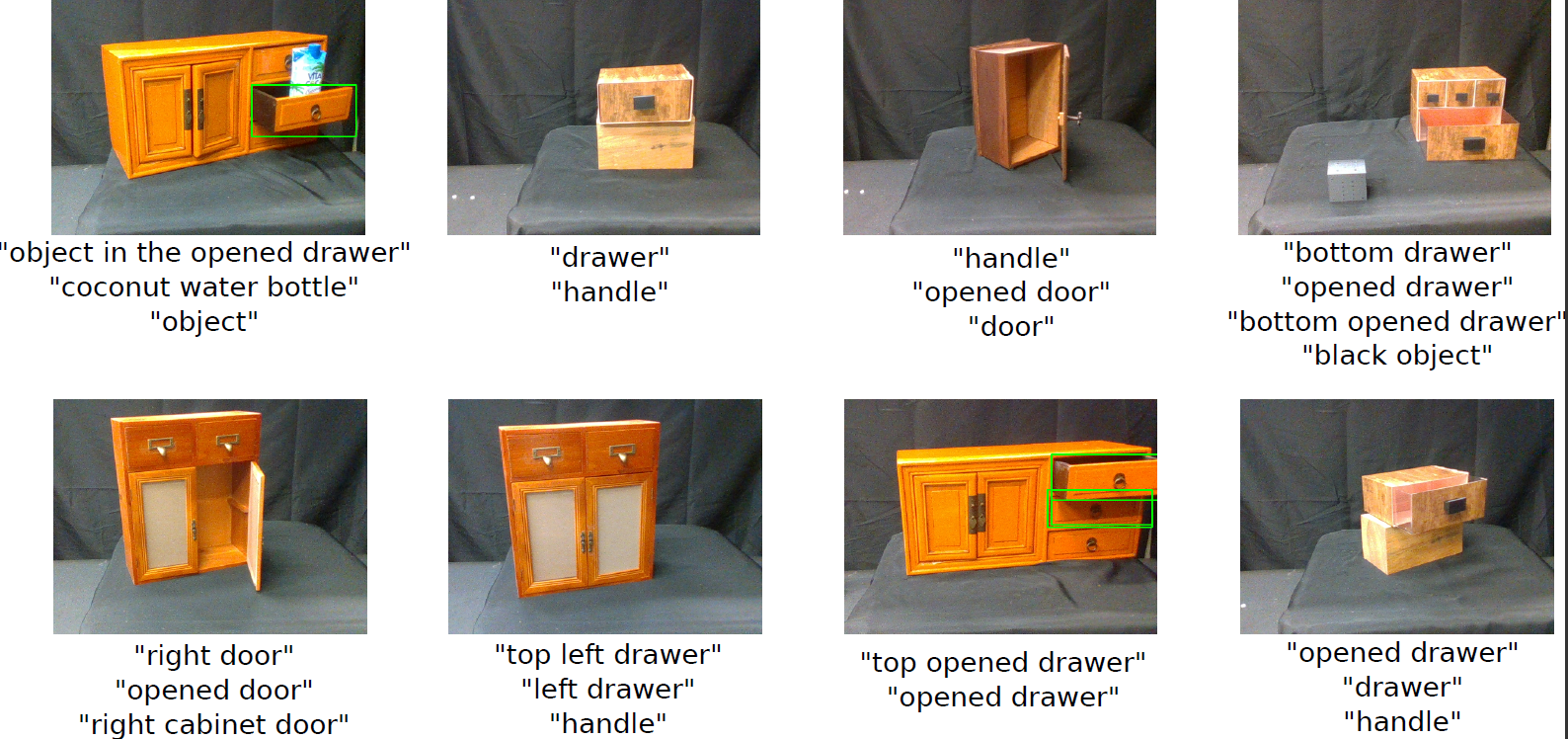}
    \caption{\footnotesize{Examples on using OWL-VIT \cite{minderer2022simple} in our robot experiments.}}
    \label{fig:owl_vit}
\end{figure}

We also visualize examples of OWL-VIT \cite{minderer2022simple} on our test observation. We tried multiple prompts and shows the best predicted results in Fig. \ref{fig:owl_vit}. Unfortunately, this approach fails to detect useful parts given the language instruction.

\begin{figure}[htb]
    \centering
    \includegraphics[width=0.45\textwidth]{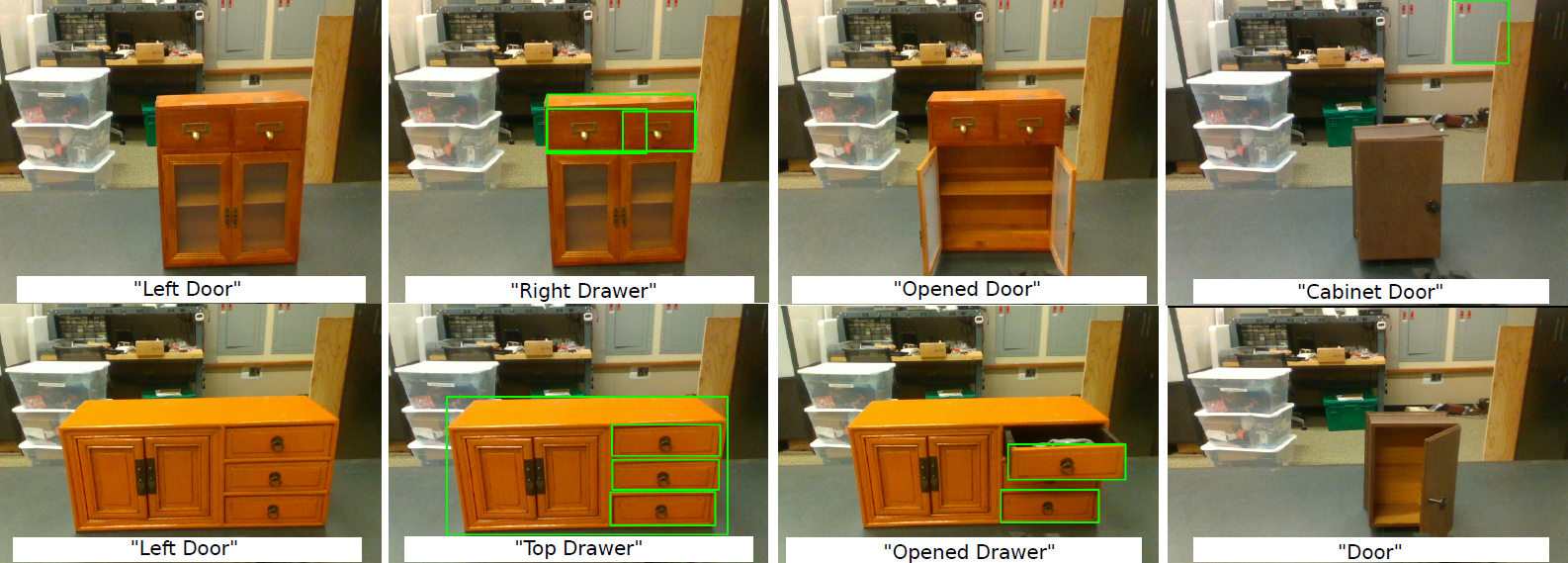}
    \caption{\footnotesize{Examples of OWL-VIT on more tasks. We observe that this method fails to detect parts that require spatial reasoning such as "top drawer" or "opened door". }}
    \label{fig:owl_vit_wild}
\end{figure}

We also provide additional visualization on more test results. We observe OWL-VIT \cite{minderer2022simple} fails on spatial reasoning and cannot provide accurate detection on specific regions or articulated parts such as "top drawer" or "opened door" as shown in Figure \ref{fig:owl_vit_wild}.

\subsection{Policy Training}
\textbf{Object-level Policy using the UR5: }
\label{sec:m2t2_appendix}
We train a language-conditioned policy for two tasks per object. We adopt network architecture from M2T2 \cite{yuan2023m2t2}, that takes RGB pointcloud and CLIP embeddings, to predict two end-effector poses, i.e where to place the gripper to grab the handle and where to release the handle to open the door.  In particular, the transformer decoder concatenates the current endeffector pose and the text features, and combines the point features from the point cloud encoder, and predicts the endeffector pose in the next step.  This model is shown in Figure \ref{fig:bc_network}.

\textbf{Visual Policy using the Stretch robot: }
We train a UNet network with extra task embedding at each encoder layer. In particular, at each layer in the encoder, the image embeddings and the task embeddings are normalized and multiplied together instead of concatenation. This is because we are only training the policy on one scene and all images are similar but task embeddings are different. The combined embeddings are connected to each corresponding decoder layer.  
To train the network, we generate an affordance dataset for the task "clean up the table surface", with predefined steps of "Open the drawer", "Pick up the object" and "Put the object in the drawer". With the advantage of a simulator, we can get the segmentation mask of each object of interest, and define a affordance map with Gaussian distribution centered at the object center. 
\begin{figure*}[h]
{\centering
    \includegraphics[width=\textwidth]{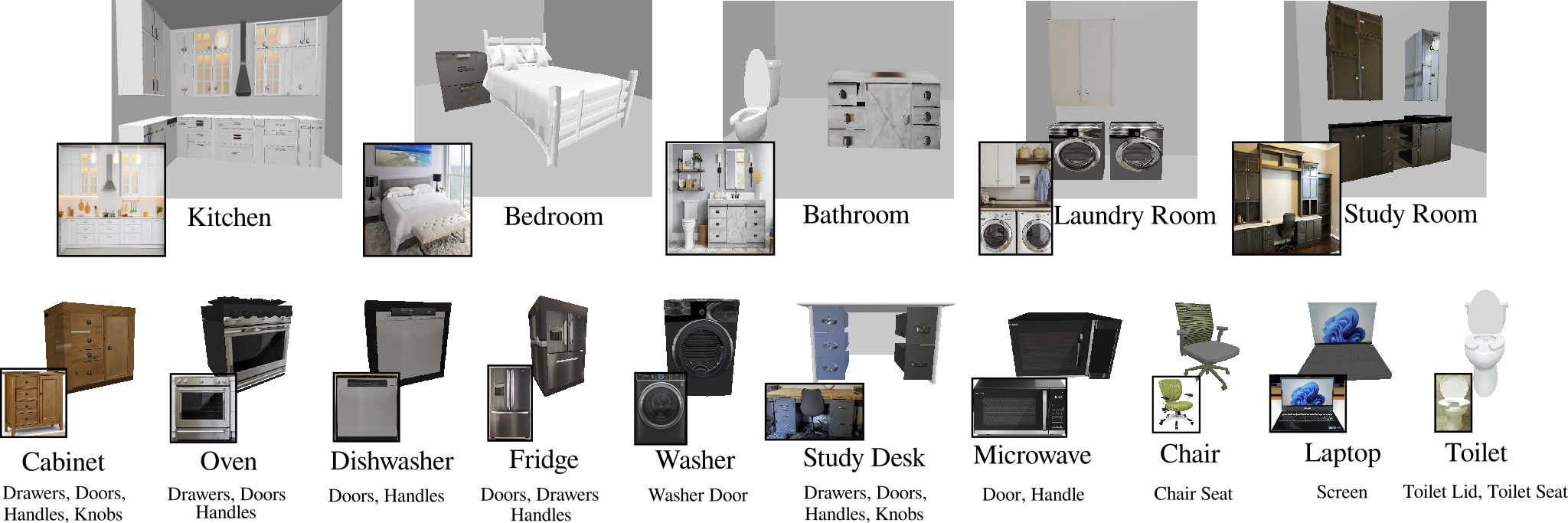}
    \captionof{figure}{A summary of categories of objects and scenes that we trained \method to predict. We visualize examples of \method predictions (large background image) given corresponding internet images (small overlaid black box). We train a single part \method for all object categories and a single global \method for all scene categories.
    Under each object category, we also list the individual articulated parts that it contains.
    See more visualizations for each new category in Section IV.}
    \label{fig:summary_figure1}
     \vspace{-0.5cm}
}
\end{figure*}
\section{Limitations and Future Work}
\label{sec:limitation_appendix}
\textbf{Part Detection}
\method relies on the performance of bounding box detection. Although the finetuned Grounding DINO improves performance than the pretrained model, there is still a gap for improvement, especially on global scene detection. It would also be interesting for future work to remove the bounding box constraints and directly predict URDFs in language format from images.

\textbf{Texture and Meshes} \method focuses on predicting kinematic URDF structures and uses predefined meshes that might not match the real-world scenes. To apply textures, we simply assume all parts are rectangular shapes, and use the bounding box of each object part to crop the image and import it into a UV map template. However, this does not work for irregular meshes such as a donut-shape door, or when the object in the image is tilted, which can be an interesting future work.

\textbf{Limited URDF Primitives}
\method currently only supports articulated objects that have limited joint types such as prismatic and revolute, and cannot predict complex objects such as cars and lamps. 

\textbf{Link Collisions} \method only predicts URDF primitives for each bounding box, which sometimes leads to a collision between two links. Further post-processing is required to resolve this issue.

\textbf{Global Scene} We observe it's challenging for object detectors such as GroundingDINO to produce high-quality part detection results. For future work, we are interested in training a better URDFormer for global scenes and expanding kitchens into other scenes such as living rooms or bathrooms. 

\textbf{Object Placement} \method only predicts the kinematic structure of the scene for a few common articulated objects, and it cannot predict poses for objects such as water bottles on the counter. In our future work, we are interested in expanding \method to cover more diverse assets and object placement. 

\textbf{Multiple Trained Components} Our pipeline contains multiple individual components that must be individually optimized.  Some of these, such as Stable Diffusion are pretrained and used off-the-shelf.  Grounding DINO also comes from a pretrained checkpoint, but was fine-tuned on new data for better performance on tasks in our desired distribution.  Finally, the global and part URDFormers were trained using our generated data.  While the use of multiple independently trained components somewhat complicates our system, it also takes the most advantage of existing pretrained models, and helps maintain consistency when constructing data via generative modeling.  Fortunately, each of these components has an unambiguous learning objective that is well-aligned with the goals of the entire system, which allows each component to be optimized in isolation.

\section{Generalization on Diverse Objects and Scenes}
\label{sec:diverse_objects_and_scenes}

\begin{figure*}[t]
    \centering
    \includegraphics[width=\textwidth]{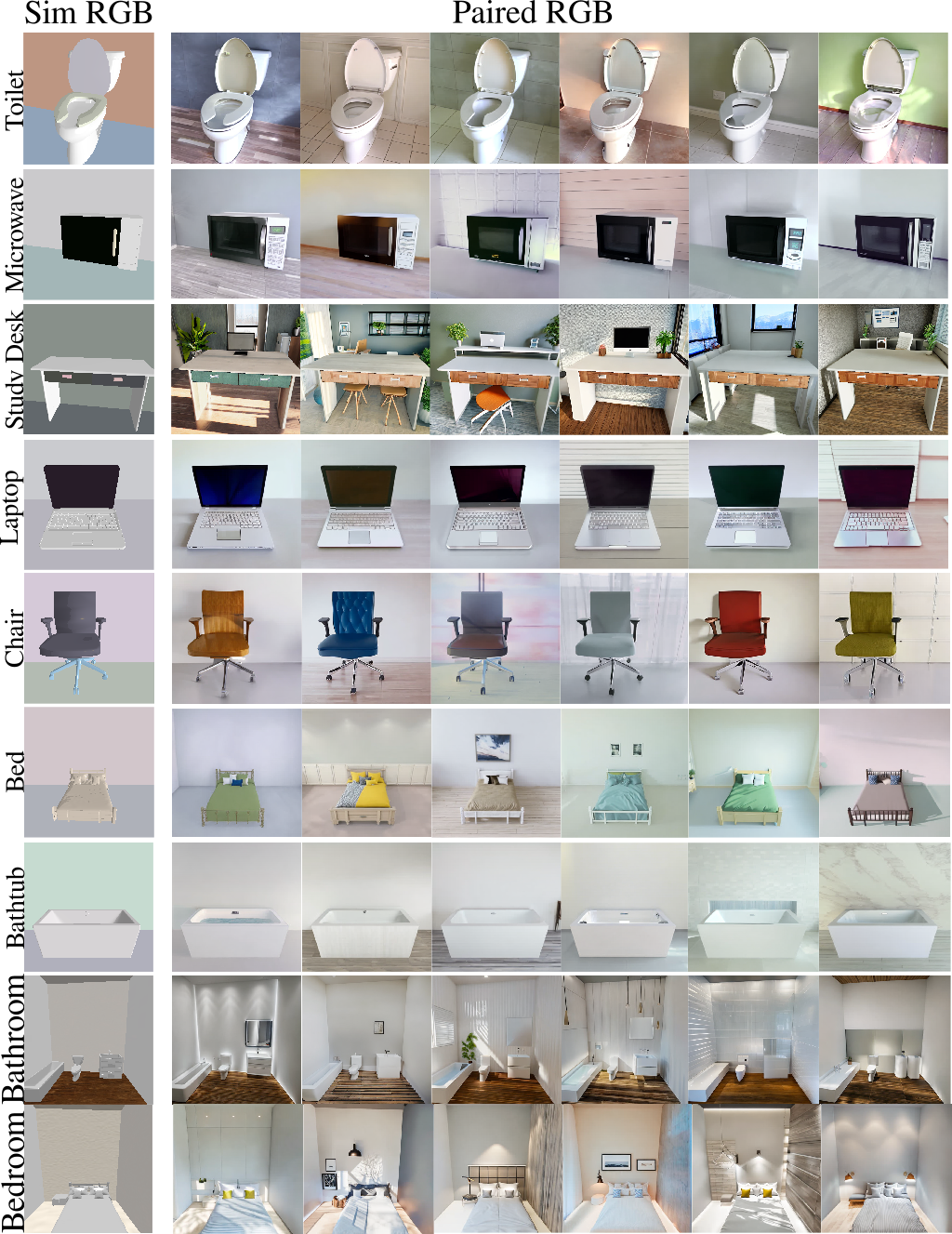}
    \caption{Visualization of diverse data generation using \method in addition to kitchen objects and scenes.}
    
    \label{fig:diverse_objects_and_scenes}
\end{figure*}

\method can be applied to a diverse set of objects and scenes. Figure \ref{fig:diverse_objects_and_scenes} visualize examples of our forward process on paired RGB images generated from synthetic RGB images. Figure \ref{fig:summary_figure1} shows a summary of categories of objects and scenes the we trained URDFormer to predict.

\clearpage
\bibliographystyle{unsrtnat}
\bibliography{main.bib}

\begin{thebibliography}{90}
\providecommand{\natexlab}[1]{#1}
\providecommand{\url}[1]{\texttt{#1}}
\expandafter\ifx\csname urlstyle\endcsname\relax
  \providecommand{\doi}[1]{doi: #1}\else
  \providecommand{\doi}{doi: \begingroup \urlstyle{rm}\Url}\fi

\bibitem[Jia et~al.(2021)Jia, Xie, Li, Chen, Zwart, Sadler, Appling, Oliver, and Read]{jia21pgml}
Xiaowei Jia, Yiqun Xie, Sheng Li, Shengyu Chen, Jacob Zwart, Jeffrey~M. Sadler, Alison~P. Appling, Samantha Oliver, and Jordan~S. Read.
\newblock Physics-guided machine learning from simulation data: An application in modeling lake and river systems.
\newblock In James Bailey, Pauli Miettinen, Yun~Sing Koh, Dacheng Tao, and Xindong Wu, editors, \emph{{IEEE} International Conference on Data Mining, {ICDM} 2021, Auckland, New Zealand, December 7-10, 2021}, pages 270--279. {IEEE}, 2021.
\newblock \doi{10.1109/ICDM51629.2021.00037}.
\newblock URL \url{https://doi.org/10.1109/ICDM51629.2021.00037}.

\bibitem[Alber et~al.(2019)Alber, Tepole, Cannon, De, Dura{-}Bernal, Garikipati, Karniadakis, Lytton, Perdikaris, Petzold, and Kuhl]{mlmodel}
Mark~S. Alber, Adrian~Buganza Tepole, William~R. Cannon, Suvranu De, Salvador Dura{-}Bernal, Krishna~C. Garikipati, George~E. Karniadakis, William~W. Lytton, Paris Perdikaris, Linda~R. Petzold, and Ellen Kuhl.
\newblock Integrating machine learning and multiscale modeling - perspectives, challenges, and opportunities in the biological, biomedical, and behavioral sciences.
\newblock \emph{npj Digit. Medicine}, 2, 2019.
\newblock \doi{10.1038/s41746-019-0193-y}.
\newblock URL \url{https://doi.org/10.1038/s41746-019-0193-y}.

\bibitem[Collins et~al.(2021)Collins, Chand, Vanderkop, and Howard]{collins21physics}
Jack Collins, Shelvin Chand, Anthony Vanderkop, and David Howard.
\newblock A review of physics simulators for robotic applications.
\newblock \emph{{IEEE} Access}, 9:\penalty0 51416--51431, 2021.
\newblock \doi{10.1109/ACCESS.2021.3068769}.
\newblock URL \url{https://doi.org/10.1109/ACCESS.2021.3068769}.

\bibitem[Narang et~al.(2022)Narang, Storey, Akinola, Macklin, Reist, Wawrzyniak, Guo, Morav{\'{a}}nszky, State, Lu, Handa, and Fox]{narang22factory}
Yashraj~S. Narang, Kier Storey, Iretiayo Akinola, Miles Macklin, Philipp Reist, Lukasz Wawrzyniak, Yunrong Guo, {\'{A}}d{\'{a}}m Morav{\'{a}}nszky, Gavriel State, Michelle Lu, Ankur Handa, and Dieter Fox.
\newblock Factory: Fast contact for robotic assembly.
\newblock In Kris Hauser, Dylan~A. Shell, and Shoudong Huang, editors, \emph{Robotics: Science and Systems XVIII, New York City, NY, USA, June 27 - July 1, 2022}, 2022.
\newblock \doi{10.15607/RSS.2022.XVIII.035}.
\newblock URL \url{https://doi.org/10.15607/RSS.2022.XVIII.035}.

\bibitem[M{\"{u}}ller et~al.(2018)M{\"{u}}ller, Casser, Lahoud, Smith, and Ghanem]{muller18sim4cv}
Matthias M{\"{u}}ller, Vincent Casser, Jean Lahoud, Neil Smith, and Bernard Ghanem.
\newblock Sim4cv: {A} photo-realistic simulator for computer vision applications.
\newblock \emph{Int. J. Comput. Vis.}, 126\penalty0 (9):\penalty0 902--919, 2018.
\newblock \doi{10.1007/s11263-018-1073-7}.
\newblock URL \url{https://doi.org/10.1007/s11263-018-1073-7}.

\bibitem[Memmel et~al.(2023{\natexlab{a}})Memmel, Bachmann, and Zamir]{memmel2023modality}
Marius Memmel, Roman Bachmann, and Amir Zamir.
\newblock Modality-invariant visual odometry for embodied vision.
\newblock In \emph{Proceedings of the IEEE/CVF Conference on Computer Vision and Pattern Recognition}, pages 21549--21559, 2023{\natexlab{a}}.
\newblock URL \url{https://openaccess.thecvf.com/content/CVPR2023/papers/Memmel_Modality-Invariant_Visual_Odometry_for_Embodied_Vision_CVPR_2023_paper.pdf}.

\bibitem[Szot et~al.(2021{\natexlab{a}})Szot, Clegg, Undersander, Wijmans, Zhao, Turner, Maestre, Mukadam, Chaplot, Maksymets, Gokaslan, Vondrus, Dharur, Meier, Galuba, Chang, Kira, Koltun, Malik, Savva, and Batra]{szot21habitat}
Andrew Szot, Alexander Clegg, Eric Undersander, Erik Wijmans, Yili Zhao, John Turner, Noah Maestre, Mustafa Mukadam, Devendra~Singh Chaplot, Oleksandr Maksymets, Aaron Gokaslan, Vladimir Vondrus, Sameer Dharur, Franziska Meier, Wojciech Galuba, Angel~X. Chang, Zsolt Kira, Vladlen Koltun, Jitendra Malik, Manolis Savva, and Dhruv Batra.
\newblock Habitat 2.0: Training home assistants to rearrange their habitat.
\newblock In Marc'Aurelio Ranzato, Alina Beygelzimer, Yann~N. Dauphin, Percy Liang, and Jennifer~Wortman Vaughan, editors, \emph{Advances in Neural Information Processing Systems 34: Annual Conference on Neural Information Processing Systems 2021, NeurIPS 2021, December 6-14, 2021, virtual}, pages 251--266, 2021{\natexlab{a}}.
\newblock URL \url{https://proceedings.neurips.cc/paper/2021/hash/021bbc7ee20b71134d53e20206bd6feb-Abstract.html}.

\bibitem[Yadav et~al.(2023)Yadav, Ramrakhya, Ramakrishnan, Gervet, Turner, Gokaslan, Maestre, Chang, Batra, Savva, Clegg, and Chaplot]{yadav23habitatmatterport}
Karmesh Yadav, Ram Ramrakhya, Santhosh~Kumar Ramakrishnan, Th{\'{e}}ophile Gervet, John Turner, Aaron Gokaslan, Noah Maestre, Angel~Xuan Chang, Dhruv Batra, Manolis Savva, Alexander~William Clegg, and Devendra~Singh Chaplot.
\newblock Habitat-matterport 3d semantics dataset.
\newblock In \emph{{IEEE/CVF} Conference on Computer Vision and Pattern Recognition, {CVPR} 2023, Vancouver, BC, Canada, June 17-24, 2023}, pages 4927--4936. {IEEE}, 2023.
\newblock \doi{10.1109/CVPR52729.2023.00477}.
\newblock URL \url{https://doi.org/10.1109/CVPR52729.2023.00477}.

\bibitem[Kolve et~al.(2017)Kolve, Mottaghi, Han, VanderBilt, Weihs, Herrasti, Deitke, Ehsani, Gordon, Zhu, et~al.]{kolve2017ai2}
Eric Kolve, Roozbeh Mottaghi, Winson Han, Eli VanderBilt, Luca Weihs, Alvaro Herrasti, Matt Deitke, Kiana Ehsani, Daniel Gordon, Yuke Zhu, et~al.
\newblock Ai2-thor: An interactive 3d environment for visual ai.
\newblock \emph{arXiv preprint arXiv:1712.05474}, 2017.
\newblock URL \url{https://arxiv.org/pdf/1712.05474.pdf}.

\bibitem[Gupta et~al.(2023)Gupta, Shepherd, and Gupta]{gupta2023predicting}
Arjun Gupta, Max Shepherd, and Saurabh Gupta.
\newblock Predicting motion plans for articulating everyday objects.
\newblock In \emph{International Conference on Robotics and Automation (ICRA)}, 2023.
\newblock URL \url{https://ieeexplore.ieee.org/document/10160752}.

\bibitem[Deitke et~al.(2022)Deitke, VanderBilt, Herrasti, Weihs, Ehsani, Salvador, Han, Kolve, Kembhavi, and Mottaghi]{deitke2022}
Matt Deitke, Eli VanderBilt, Alvaro Herrasti, Luca Weihs, Kiana Ehsani, Jordi Salvador, Winson Han, Eric Kolve, Aniruddha Kembhavi, and Roozbeh Mottaghi.
\newblock Procthor: Large-scale embodied ai using procedural generation.
\newblock \emph{Advances in Neural Information Processing Systems}, 35:\penalty0 5982--5994, 2022.
\newblock URL \url{https://arxiv.org/pdf/2206.06994.pdf}.

\bibitem[Poole et~al.(2023)Poole, Jain, Barron, and Mildenhall]{dreamfusion}
Ben Poole, Ajay Jain, Jonathan~T. Barron, and Ben Mildenhall.
\newblock Dreamfusion: Text-to-3d using 2d diffusion.
\newblock In \emph{The Eleventh International Conference on Learning Representations, {ICLR} 2023, Kigali, Rwanda, May 1-5, 2023}. OpenReview.net, 2023.
\newblock URL \url{https://openreview.net/pdf?id=FjNys5c7VyY}.

\bibitem[Wang et~al.(2023{\natexlab{a}})Wang, Xian, Chen, Wang, Wang, Fragkiadaki, Erickson, Held, and Gan]{wang2023robogen}
Yufei Wang, Zhou Xian, Feng Chen, Tsun-Hsuan Wang, Yian Wang, Katerina Fragkiadaki, Zackory Erickson, David Held, and Chuang Gan.
\newblock Robogen: Towards unleashing infinite data for automated robot learning via generative simulation.
\newblock \emph{arXiv preprint arXiv:2311.01455}, 2023{\natexlab{a}}.
\newblock URL \url{https://arxiv.org/pdf/2311.01455.pdf}.

\bibitem[Wang et~al.(2023{\natexlab{b}})Wang, Ling, Yuan, Shridhar, Bao, Qin, Wang, Xu, and Wang]{wang2023gen}
Lirui Wang, Yiyang Ling, Zhecheng Yuan, Mohit Shridhar, Chen Bao, Yuzhe Qin, Bailin Wang, Huazhe Xu, and Xiaolong Wang.
\newblock Gensim: Generating robotic simulation tasks via large language models.
\newblock In \emph{Arxiv}, 2023{\natexlab{b}}.
\newblock URL \url{https://arxiv.org/pdf/2310.01361.pdf}.

\bibitem[Liu et~al.(2023{\natexlab{a}})Liu, Xu, Jin, Chen, Xu, Su, et~al.]{liu2023one}
Minghua Liu, Chao Xu, Haian Jin, Linghao Chen, Zexiang Xu, Hao Su, et~al.
\newblock One-2-3-45: Any single image to 3d mesh in 45 seconds without per-shape optimization.
\newblock \emph{arXiv preprint arXiv:2306.16928}, 2023{\natexlab{a}}.
\newblock URL \url{https://arxiv.org/pdf/2306.16928.pdf}.

\bibitem[Fang et~al.(2023)Fang, Hu, Luo, and Tan]{fang2023ctrl}
Chuan Fang, Xiaotao Hu, Kunming Luo, and Ping Tan.
\newblock Ctrl-room: Controllable text-to-3d room meshes generation with layout constraints.
\newblock \emph{arXiv preprint arXiv:2310.03602}, 2023.
\newblock URL \url{https://arxiv.org/pdf/2310.03602.pdf}.

\bibitem[Raistrick et~al.(2023)Raistrick, Lipson, Ma, Mei, Wang, Zuo, Kayan, Wen, Han, Wang, Newell, Law, Goyal, Yang, and Deng]{raistrick23procedural}
Alexander Raistrick, Lahav Lipson, Zeyu Ma, Lingjie Mei, Mingzhe Wang, Yiming Zuo, Karhan Kayan, Hongyu Wen, Beining Han, Yihan Wang, Alejandro Newell, Hei Law, Ankit Goyal, Kaiyu Yang, and Jia Deng.
\newblock Infinite photorealistic worlds using procedural generation.
\newblock In \emph{{IEEE/CVF} Conference on Computer Vision and Pattern Recognition, {CVPR} 2023, Vancouver, BC, Canada, June 17-24, 2023}, pages 12630--12641. {IEEE}, 2023.
\newblock \doi{10.1109/CVPR52729.2023.01215}.
\newblock URL \url{https://doi.org/10.1109/CVPR52729.2023.01215}.

\bibitem[Rombach et~al.(2022)Rombach, Blattmann, Lorenz, Esser, and Ommer]{rombach2022high}
Robin Rombach, Andreas Blattmann, Dominik Lorenz, Patrick Esser, and Bj{\"o}rn Ommer.
\newblock High-resolution image synthesis with latent diffusion models.
\newblock In \emph{Proceedings of the IEEE/CVF conference on computer vision and pattern recognition}, pages 10684--10695, 2022.
\newblock URL \url{https://openaccess.thecvf.com/content/CVPR2022/papers/Rombach_High-Resolution_Image_Synthesis_With_Latent_Diffusion_Models_CVPR_2022_paper.pdf}.

\bibitem[Mo et~al.(2019)Mo, Zhu, Chang, Yi, Tripathi, Guibas, and Su]{mopartnet}
Kaichun Mo, Shilin Zhu, Angel~X. Chang, Li~Yi, Subarna Tripathi, Leonidas~J. Guibas, and Hao Su.
\newblock Partnet: {A} large-scale benchmark for fine-grained and hierarchical part-level 3d object understanding.
\newblock In \emph{{IEEE} Conference on Computer Vision and Pattern Recognition, {CVPR} 2019, Long Beach, CA, USA, June 16-20, 2019}, pages 909--918. Computer Vision Foundation / {IEEE}, 2019.
\newblock \doi{10.1109/CVPR.2019.00100}.
\newblock URL \url{http://openaccess.thecvf.com/content\_CVPR\_2019/html/Mo\_PartNet\_A\_Large-Scale\_Benchmark\_for\_Fine-Grained\_and\_Hierarchical\_Part-Level\_3D\_CVPR\_2019\_paper.html}.

\bibitem[Dosovitskiy et~al.(2020)Dosovitskiy, Beyer, Kolesnikov, Weissenborn, Zhai, Unterthiner, Dehghani, Minderer, Heigold, Gelly, et~al.]{dosovitskiy2020image}
Alexey Dosovitskiy, Lucas Beyer, Alexander Kolesnikov, Dirk Weissenborn, Xiaohua Zhai, Thomas Unterthiner, Mostafa Dehghani, Matthias Minderer, Georg Heigold, Sylvain Gelly, et~al.
\newblock An image is worth 16x16 words: Transformers for image recognition at scale.
\newblock \emph{arXiv preprint arXiv:2010.11929}, 2020.
\newblock URL \url{https://arxiv.org/pdf/2010.11929.pdf}.

\bibitem[He et~al.(2017)He, Gkioxari, Doll{\'a}r, and Girshick]{he2017mask}
Kaiming He, Georgia Gkioxari, Piotr Doll{\'a}r, and Ross Girshick.
\newblock Mask r-cnn.
\newblock In \emph{Proceedings of the IEEE international conference on computer vision}, pages 2961--2969, 2017.
\newblock URL \url{https://arxiv.org/pdf/1703.06870.pdf}.

\bibitem[Vaswani et~al.(2017)Vaswani, Shazeer, Parmar, Uszkoreit, Jones, Gomez, Kaiser, and Polosukhin]{vaswani2017attention}
Ashish Vaswani, Noam Shazeer, Niki Parmar, Jakob Uszkoreit, Llion Jones, Aidan~N Gomez, {\L}ukasz Kaiser, and Illia Polosukhin.
\newblock Attention is all you need.
\newblock \emph{Advances in neural information processing systems}, 30, 2017.
\newblock URL \url{https://proceedings.neurips.cc/paper_files/paper/2017/file/3f5ee243547dee91fbd053c1c4a845aa-Paper.pdf}.

\bibitem[Yang et~al.(2023{\natexlab{a}})Yang, Peng, Li, Guo, Chen, Li, Ma, Zhou, Zhang, Loy, and Liu]{yang2023pvsg}
Jingkang Yang, Wenxuan Peng, Xiangtai Li, Zujin Guo, Liangyu Chen, Bo~Li, Zheng Ma, Kaiyang Zhou, Wayne Zhang, Chen~Change Loy, and Ziwei Liu.
\newblock Panoptic video scene graph generation.
\newblock In \emph{CVPR}, 2023{\natexlab{a}}.
\newblock URL \url{https://openaccess.thecvf.com/content/CVPR2023/papers/Yang_Panoptic_Video_Scene_Graph_Generation_CVPR_2023_paper.pdf}.

\bibitem[Sundaralingam et~al.(2023)Sundaralingam, Hari, Fishman, Garrett, Wyk, Blukis, Millane, Oleynikova, Handa, Ramos, Ratliff, and Fox]{curobo_report23}
Balakumar Sundaralingam, Siva Kumar~Sastry Hari, Adam Fishman, Caelan Garrett, Karl~Van Wyk, Valts Blukis, Alexander Millane, Helen Oleynikova, Ankur Handa, Fabio Ramos, Nathan Ratliff, and Dieter Fox.
\newblock curobo: Parallelized collision-free minimum-jerk robot motion generation, 2023.
\newblock URL \url{https://arxiv.org/pdf/2310.17274.pdf}.

\bibitem[Yuan et~al.(2023)Yuan, Murali, Mousavian, and Fox]{yuan2023m2t2}
Wentao Yuan, Adithyavairavan Murali, Arsalan Mousavian, and Dieter Fox.
\newblock M2t2: Multi-task masked transformer for object-centric pick and place.
\newblock In \emph{7th Annual Conference on Robot Learning}, 2023.
\newblock URL \url{https://arxiv.org/pdf/2311.00926.pdf}.

\bibitem[Memmel et~al.(2022)Memmel, Liu, Tateo, and Peters]{memmel2022dimensionality}
Marius Memmel, Puze Liu, Davide Tateo, and Jan Peters.
\newblock Dimensionality reduction and prioritized exploration for policy search.
\newblock In \emph{International Conference on Artificial Intelligence and Statistics}, pages 2134--2157. PMLR, 2022.
\newblock URL \url{https://proceedings.mlr.press/v151/memmel22a.html}.

\bibitem[Schulman et~al.(2017)Schulman, Wolski, Dhariwal, Radford, and Klimov]{schulman2017proximal}
John Schulman, Filip Wolski, Prafulla Dhariwal, Alec Radford, and Oleg Klimov.
\newblock Proximal policy optimization algorithms.
\newblock \emph{arXiv preprint arXiv:1707.06347}, 2017.
\newblock URL \url{https://arxiv.org/pdf/1707.06347.pdf}.

\bibitem[Haarnoja et~al.(2018)Haarnoja, Zhou, Abbeel, and Levine]{haarnojasac}
Tuomas Haarnoja, Aurick Zhou, Pieter Abbeel, and Sergey Levine.
\newblock Soft actor-critic: Off-policy maximum entropy deep reinforcement learning with a stochastic actor.
\newblock In Jennifer~G. Dy and Andreas Krause, editors, \emph{Proceedings of the 35th International Conference on Machine Learning, {ICML} 2018, Stockholmsm{\"{a}}ssan, Stockholm, Sweden, July 10-15, 2018}, volume~80 of \emph{Proceedings of Machine Learning Research}, pages 1856--1865. {PMLR}, 2018.
\newblock URL \url{http://proceedings.mlr.press/v80/haarnoja18b.html}.

\bibitem[Minderer et~al.(2022)Minderer, Gritsenko, Stone, Neumann, Weissenborn, Dosovitskiy, Mahendran, Arnab, Dehghani, Shen, et~al.]{minderer2022simple}
Matthias Minderer, Alexey Gritsenko, Austin Stone, Maxim Neumann, Dirk Weissenborn, Alexey Dosovitskiy, Aravindh Mahendran, Anurag Arnab, Mostafa Dehghani, Zhuoran Shen, et~al.
\newblock Simple open-vocabulary object detection.
\newblock In \emph{European Conference on Computer Vision}, 2022.
\newblock URL \url{https://www.ecva.net/papers/eccv_2022/papers_ECCV/papers/136700714.pdf}.

\bibitem[Collaboration et~al.(2023)Collaboration, Padalkar, Pooley, Jain, Bewley, Herzog, Irpan, Khazatsky, Rai, Singh, Brohan, Raffin, Wahid, Burgess-Limerick, Kim, Schölkopf, Ichter, Lu, Xu, Finn, Xu, Chi, Huang, Chan, Pan, Fu, Devin, Driess, Pathak, Shah, Büchler, Kalashnikov, Sadigh, Johns, Ceola, Xia, Stulp, Zhou, Sukhatme, Salhotra, Yan, Schiavi, Su, Fang, Shi, Amor, Christensen, Furuta, Walke, Fang, Mordatch, Radosavovic, Leal, Liang, Kim, Schneider, Hsu, Bohg, Bingham, Wu, Wu, Luo, Gu, Tan, Oh, Malik, Tompson, Yang, Lim, Silvério, Han, Rao, Pertsch, Hausman, Go, Gopalakrishnan, Goldberg, Byrne, Oslund, Kawaharazuka, Zhang, Majd, Rana, Srinivasan, Chen, Pinto, Tan, Ott, Lee, Tomizuka, Du, Ahn, Zhang, Ding, Srirama, Sharma, Kim, Kanazawa, Hansen, Heess, Joshi, Suenderhauf, Palo, Shafiullah, Mees, Kroemer, Sanketi, Wohlhart, Xu, Sermanet, Sundaresan, Vuong, Rafailov, Tian, Doshi, Martín-Martín, Mendonca, Shah, Hoque, Julian, Bustamante, Kirmani, Levine, Moore, Bahl, Dass, Song, Xu, Haldar, Adebola,
  Guist, Nasiriany, Schaal, Welker, Tian, Dasari, Belkhale, Osa, Harada, Matsushima, Xiao, Yu, Ding, Davchev, Zhao, Armstrong, Darrell, Jain, Vanhoucke, Zhan, Zhou, Burgard, Chen, Wang, Zhu, Li, Lu, Chebotar, Zhou, Zhu, Xu, Wang, Bisk, Cho, Lee, Cui, hua Wu, Tang, Zhu, Li, Iwasawa, Matsuo, Xu, and Cui]{open_x_embodiment_rt_x_2023}
Open X-Embodiment Collaboration, Abhishek Padalkar, Acorn Pooley, Ajinkya Jain, Alex Bewley, Alex Herzog, Alex Irpan, Alexander Khazatsky, Anant Rai, Anikait Singh, Anthony Brohan, Antonin Raffin, Ayzaan Wahid, Ben Burgess-Limerick, Beomjoon Kim, Bernhard Schölkopf, Brian Ichter, Cewu Lu, Charles Xu, Chelsea Finn, Chenfeng Xu, Cheng Chi, Chenguang Huang, Christine Chan, Chuer Pan, Chuyuan Fu, Coline Devin, Danny Driess, Deepak Pathak, Dhruv Shah, Dieter Büchler, Dmitry Kalashnikov, Dorsa Sadigh, Edward Johns, Federico Ceola, Fei Xia, Freek Stulp, Gaoyue Zhou, Gaurav~S. Sukhatme, Gautam Salhotra, Ge~Yan, Giulio Schiavi, Hao Su, Hao-Shu Fang, Haochen Shi, Heni~Ben Amor, Henrik~I Christensen, Hiroki Furuta, Homer Walke, Hongjie Fang, Igor Mordatch, Ilija Radosavovic, Isabel Leal, Jacky Liang, Jaehyung Kim, Jan Schneider, Jasmine Hsu, Jeannette Bohg, Jeffrey Bingham, Jiajun Wu, Jialin Wu, Jianlan Luo, Jiayuan Gu, Jie Tan, Jihoon Oh, Jitendra Malik, Jonathan Tompson, Jonathan Yang, Joseph~J. Lim, João
  Silvério, Junhyek Han, Kanishka Rao, Karl Pertsch, Karol Hausman, Keegan Go, Keerthana Gopalakrishnan, Ken Goldberg, Kendra Byrne, Kenneth Oslund, Kento Kawaharazuka, Kevin Zhang, Keyvan Majd, Krishan Rana, Krishnan Srinivasan, Lawrence~Yunliang Chen, Lerrel Pinto, Liam Tan, Lionel Ott, Lisa Lee, Masayoshi Tomizuka, Maximilian Du, Michael Ahn, Mingtong Zhang, Mingyu Ding, Mohan~Kumar Srirama, Mohit Sharma, Moo~Jin Kim, Naoaki Kanazawa, Nicklas Hansen, Nicolas Heess, Nikhil~J Joshi, Niko Suenderhauf, Norman~Di Palo, Nur Muhammad~Mahi Shafiullah, Oier Mees, Oliver Kroemer, Pannag~R Sanketi, Paul Wohlhart, Peng Xu, Pierre Sermanet, Priya Sundaresan, Quan Vuong, Rafael Rafailov, Ran Tian, Ria Doshi, Roberto Martín-Martín, Russell Mendonca, Rutav Shah, Ryan Hoque, Ryan Julian, Samuel Bustamante, Sean Kirmani, Sergey Levine, Sherry Moore, Shikhar Bahl, Shivin Dass, Shuran Song, Sichun Xu, Siddhant Haldar, Simeon Adebola, Simon Guist, Soroush Nasiriany, Stefan Schaal, Stefan Welker, Stephen Tian, Sudeep Dasari,
  Suneel Belkhale, Takayuki Osa, Tatsuya Harada, Tatsuya Matsushima, Ted Xiao, Tianhe Yu, Tianli Ding, Todor Davchev, Tony~Z. Zhao, Travis Armstrong, Trevor Darrell, Vidhi Jain, Vincent Vanhoucke, Wei Zhan, Wenxuan Zhou, Wolfram Burgard, Xi~Chen, Xiaolong Wang, Xinghao Zhu, Xuanlin Li, Yao Lu, Yevgen Chebotar, Yifan Zhou, Yifeng Zhu, Ying Xu, Yixuan Wang, Yonatan Bisk, Yoonyoung Cho, Youngwoon Lee, Yuchen Cui, Yueh hua Wu, Yujin Tang, Yuke Zhu, Yunzhu Li, Yusuke Iwasawa, Yutaka Matsuo, Zhuo Xu, and Zichen~Jeff Cui.
\newblock Open {X-E}mbodiment: Robotic learning datasets and {RT-X} models.
\newblock \url{https://arxiv.org/abs/2310.08864}, 2023.

\bibitem[Gupta et~al.(2021)Gupta, Yu, Zhao, Kumar, Rovinsky, Xu, Devlin, and Levine]{gupta21mtrf}
Abhishek Gupta, Justin Yu, Tony~Z. Zhao, Vikash Kumar, Aaron Rovinsky, Kelvin Xu, Thomas Devlin, and Sergey Levine.
\newblock Reset-free reinforcement learning via multi-task learning: Learning dexterous manipulation behaviors without human intervention.
\newblock In \emph{{IEEE} International Conference on Robotics and Automation, {ICRA} 2021, Xi'an, China, May 30 - June 5, 2021}, pages 6664--6671. {IEEE}, 2021.
\newblock \doi{10.1109/ICRA48506.2021.9561384}.
\newblock URL \url{https://doi.org/10.1109/ICRA48506.2021.9561384}.

\bibitem[Liu et~al.(2023{\natexlab{b}})Liu, Zeng, Ren, Li, Zhang, Yang, Li, Yang, Su, Zhu, et~al.]{liu2023grounding}
Shilong Liu, Zhaoyang Zeng, Tianhe Ren, Feng Li, Hao Zhang, Jie Yang, Chunyuan Li, Jianwei Yang, Hang Su, Jun Zhu, et~al.
\newblock Grounding dino: Marrying dino with grounded pre-training for open-set object detection.
\newblock \emph{arXiv preprint arXiv:2303.05499}, 2023{\natexlab{b}}.
\newblock URL \url{https://arxiv.org/pdf/2303.05499.pdf}.

\bibitem[Wortsman et~al.(2022)Wortsman, Ilharco, Gadre, Roelofs, Gontijo-Lopes, Morcos, Namkoong, Farhadi, Carmon, Kornblith, et~al.]{wortsman2022model}
Mitchell Wortsman, Gabriel Ilharco, Samir~Ya Gadre, Rebecca Roelofs, Raphael Gontijo-Lopes, Ari~S Morcos, Hongseok Namkoong, Ali Farhadi, Yair Carmon, Simon Kornblith, et~al.
\newblock Model soups: averaging weights of multiple fine-tuned models improves accuracy without increasing inference time.
\newblock In \emph{International Conference on Machine Learning}, pages 23965--23998. PMLR, 2022.

\bibitem[Coumans and Bai(2016--2021)]{coumans2021}
Erwin Coumans and Yunfei Bai.
\newblock Pybullet, a python module for physics simulation for games, robotics and machine learning.
\newblock \url{http://pybullet.org}, 2016--2021.

\bibitem[Huang et~al.(2023)Huang, Wang, Zhang, Li, Wu, and Fei-Fei]{huang2023voxposer}
Wenlong Huang, Chen Wang, Ruohan Zhang, Yunzhu Li, Jiajun Wu, and Li~Fei-Fei.
\newblock Voxposer: Composable 3d value maps for robotic manipulation with language models.
\newblock \emph{arXiv preprint arXiv:2307.05973}, 2023.

\bibitem[Tobin et~al.(2017)Tobin, Fong, Ray, Schneider, Zaremba, and Abbeel]{tobin17domainrand}
Josh Tobin, Rachel Fong, Alex Ray, Jonas Schneider, Wojciech Zaremba, and Pieter Abbeel.
\newblock Domain randomization for transferring deep neural networks from simulation to the real world.
\newblock In \emph{2017 {IEEE/RSJ} International Conference on Intelligent Robots and Systems, {IROS} 2017, Vancouver, BC, Canada, September 24-28, 2017}, pages 23--30. {IEEE}, 2017.
\newblock \doi{10.1109/IROS.2017.8202133}.
\newblock URL \url{https://doi.org/10.1109/IROS.2017.8202133}.

\bibitem[Besl and McKay(1992)]{besl1992method}
Paul~J Besl and Neil~D McKay.
\newblock Method for registration of 3-d shapes.
\newblock In \emph{Sensor fusion IV: control paradigms and data structures}, volume 1611, pages 586--606. Spie, 1992.
\newblock URL \url{https://www.researchgate.net/publication/3191994_A_method_for_registration_of_3-D_shapes_IEEE_Trans_Pattern_Anal_Mach_Intell}.

\bibitem[Cimpoi et~al.(2014)Cimpoi, Maji, Kokkinos, Mohamed, , and Vedaldi]{cimpoi14describing}
M.~Cimpoi, S.~Maji, I.~Kokkinos, S.~Mohamed, , and A.~Vedaldi.
\newblock Describing textures in the wild.
\newblock In \emph{Proceedings of the {IEEE} Conf. on Computer Vision and Pattern Recognition ({CVPR})}, 2014.
\newblock URL \url{https://arxiv.org/pdf/1311.3618.pdf}.

\bibitem[Izadi et~al.(2011)Izadi, Kim, Hilliges, Molyneaux, Newcombe, Kohli, Shotton, Hodges, Freeman, Davison, et~al.]{izadi2011kinectfusion}
Shahram Izadi, David Kim, Otmar Hilliges, David Molyneaux, Richard Newcombe, Pushmeet Kohli, Jamie Shotton, Steve Hodges, Dustin Freeman, Andrew Davison, et~al.
\newblock Kinectfusion: real-time 3d reconstruction and interaction using a moving depth camera.
\newblock In \emph{Proceedings of the 24th annual ACM symposium on User interface software and technology}, pages 559--568, 2011.

\bibitem[Agarwal et~al.(2011)Agarwal, Furukawa, Snavely, Simon, Curless, Seitz, and Szeliski]{agarwal2011building}
Sameer Agarwal, Yasutaka Furukawa, Noah Snavely, Ian Simon, Brian Curless, Steven~M Seitz, and Richard Szeliski.
\newblock Building rome in a day.
\newblock \emph{Communications of the ACM}, 54\penalty0 (10):\penalty0 105--112, 2011.

\bibitem[Henry et~al.(2012)Henry, Krainin, Herbst, Ren, and Fox]{henry2012rgb}
Peter Henry, Michael Krainin, Evan Herbst, Xiaofeng Ren, and Dieter Fox.
\newblock Rgb-d mapping: Using kinect-style depth cameras for dense 3d modeling of indoor environments.
\newblock \emph{The international journal of Robotics Research}, 31\penalty0 (5):\penalty0 647--663, 2012.

\bibitem[Mur-Artal et~al.(2015)Mur-Artal, Montiel, and Tardos]{mur2015orb}
Raul Mur-Artal, Jose Maria~Martinez Montiel, and Juan~D Tardos.
\newblock Orb-slam: a versatile and accurate monocular slam system.
\newblock \emph{IEEE transactions on robotics}, 31\penalty0 (5):\penalty0 1147--1163, 2015.

\bibitem[Park et~al.(2019)Park, Florence, Straub, Newcombe, and Lovegrove]{park2019deepsdf}
Jeong~Joon Park, Peter Florence, Julian Straub, Richard Newcombe, and Steven Lovegrove.
\newblock Deepsdf: Learning continuous signed distance functions for shape representation.
\newblock In \emph{Proceedings of the IEEE/CVF conference on computer vision and pattern recognition}, pages 165--174, 2019.

\bibitem[Mildenhall et~al.(2021)Mildenhall, Srinivasan, Tancik, Barron, Ramamoorthi, and Ng]{mildenhall2021nerf}
Ben Mildenhall, Pratul~P Srinivasan, Matthew Tancik, Jonathan~T Barron, Ravi Ramamoorthi, and Ren Ng.
\newblock Nerf: Representing scenes as neural radiance fields for view synthesis.
\newblock \emph{Communications of the ACM}, 65\penalty0 (1):\penalty0 99--106, 2021.

\bibitem[Kerbl et~al.(2023)Kerbl, Kopanas, Leimk{\"u}hler, and Drettakis]{kerbl20233d}
Bernhard Kerbl, Georgios Kopanas, Thomas Leimk{\"u}hler, and George Drettakis.
\newblock 3d gaussian splatting for real-time radiance field rendering.
\newblock \emph{ACM Transactions on Graphics (ToG)}, 42\penalty0 (4):\penalty0 1--14, 2023.

\bibitem[Xiang et~al.(2020)Xiang, Qin, Mo, Xia, Zhu, Liu, Liu, Jiang, Yuan, Wang, et~al.]{xiang2020sapien}
Fanbo Xiang, Yuzhe Qin, Kaichun Mo, Yikuan Xia, Hao Zhu, Fangchen Liu, Minghua Liu, Hanxiao Jiang, Yifu Yuan, He~Wang, et~al.
\newblock Sapien: A simulated part-based interactive environment.
\newblock In \emph{Proceedings of the IEEE/CVF Conference on Computer Vision and Pattern Recognition}, pages 11097--11107, 2020.
\newblock URL \url{https://openaccess.thecvf.com/content_CVPR_2020/papers/Xiang_SAPIEN_A_SimulAted_Part-Based_Interactive_ENvironment_CVPR_2020_paper.pdf}.

\bibitem[Szot et~al.(2021{\natexlab{b}})Szot, Clegg, Undersander, Wijmans, Zhao, Turner, Maestre, Mukadam, Chaplot, Maksymets, et~al.]{szot2021habitat}
Andrew Szot, Alexander Clegg, Eric Undersander, Erik Wijmans, Yili Zhao, John Turner, Noah Maestre, Mustafa Mukadam, Devendra~Singh Chaplot, Oleksandr Maksymets, et~al.
\newblock Habitat 2.0: Training home assistants to rearrange their habitat.
\newblock \emph{Advances in Neural Information Processing Systems}, 34:\penalty0 251--266, 2021{\natexlab{b}}.

\bibitem[Deitke et~al.(2023{\natexlab{a}})Deitke, Schwenk, Salvador, Weihs, Michel, VanderBilt, Schmidt, Ehsani, Kembhavi, and Farhadi]{deitke2023objaverse}
Matt Deitke, Dustin Schwenk, Jordi Salvador, Luca Weihs, Oscar Michel, Eli VanderBilt, Ludwig Schmidt, Kiana Ehsani, Aniruddha Kembhavi, and Ali Farhadi.
\newblock Objaverse: A universe of annotated 3d objects.
\newblock In \emph{Proceedings of the IEEE/CVF Conference on Computer Vision and Pattern Recognition}, pages 13142--13153, 2023{\natexlab{a}}.

\bibitem[Yang et~al.(2023{\natexlab{b}})Yang, Sun, Weihs, VanderBilt, Herrasti, Han, Wu, Haber, Krishna, Liu, et~al.]{yang2023holodeck}
Yue Yang, Fan-Yun Sun, Luca Weihs, Eli VanderBilt, Alvaro Herrasti, Winson Han, Jiajun Wu, Nick Haber, Ranjay Krishna, Lingjie Liu, et~al.
\newblock Holodeck: Language guided generation of 3d embodied ai environments.
\newblock \emph{arXiv preprint arXiv:2312.09067}, 2023{\natexlab{b}}.

\bibitem[Mart{\'\i}n-Mart{\'\i}n et~al.(2019)Mart{\'\i}n-Mart{\'\i}n, Eppner, and Brock]{martin2019rbo}
Roberto Mart{\'\i}n-Mart{\'\i}n, Clemens Eppner, and Oliver Brock.
\newblock The rbo dataset of articulated objects and interactions.
\newblock \emph{The International Journal of Robotics Research}, 38\penalty0 (9):\penalty0 1013--1019, 2019.

\bibitem[Xia et~al.(2020)Xia, Shen, Li, Kasimbeg, Tchapmi, Toshev, Mart{\'\i}n-Mart{\'\i}n, and Savarese]{xia2020interactive}
Fei Xia, William~B Shen, Chengshu Li, Priya Kasimbeg, Micael~Edmond Tchapmi, Alexander Toshev, Roberto Mart{\'\i}n-Mart{\'\i}n, and Silvio Savarese.
\newblock Interactive gibson benchmark: A benchmark for interactive navigation in cluttered environments.
\newblock \emph{IEEE Robotics and Automation Letters}, 5\penalty0 (2):\penalty0 713--720, 2020.

\bibitem[Liu et~al.(2022)Liu, Xu, Fu, Qian, Yu, Han, and Lu]{liu2022akb}
Liu Liu, Wenqiang Xu, Haoyuan Fu, Sucheng Qian, Qiaojun Yu, Yang Han, and Cewu Lu.
\newblock Akb-48: A real-world articulated object knowledge base.
\newblock In \emph{Proceedings of the IEEE/CVF Conference on Computer Vision and Pattern Recognition}, pages 14809--14818, 2022.

\bibitem[Yang et~al.(2021)Yang, Sun, Jampani, Vlasic, Cole, Chang, Ramanan, Freeman, and Liu]{yang2021lasr}
Gengshan Yang, Deqing Sun, Varun Jampani, Daniel Vlasic, Forrester Cole, Huiwen Chang, Deva Ramanan, William~T Freeman, and Ce~Liu.
\newblock Lasr: Learning articulated shape reconstruction from a monocular video.
\newblock In \emph{Proceedings of the IEEE/CVF Conference on Computer Vision and Pattern Recognition}, pages 15980--15989, 2021.

\bibitem[Qian et~al.(2022)Qian, Jin, Rockwell, Chen, and Fouhey]{Qian22}
Shengyi Qian, Linyi Jin, Chris Rockwell, Siyi Chen, and David~F. Fouhey.
\newblock Understanding 3d object articulation in internet videos.
\newblock In \emph{CVPR}, 2022.

\bibitem[Wei et~al.(2022)Wei, Chabra, Ma, Lassner, Zollh{\"o}fer, Rusinkiewicz, Sweeney, Newcombe, and Slavcheva]{wei2022self}
Fangyin Wei, Rohan Chabra, Lingni Ma, Christoph Lassner, Michael Zollh{\"o}fer, Szymon Rusinkiewicz, Chris Sweeney, Richard Newcombe, and Mira Slavcheva.
\newblock Self-supervised neural articulated shape and appearance models.
\newblock In \emph{Proceedings of the IEEE/CVF Conference on Computer Vision and Pattern Recognition}, pages 15816--15826, 2022.

\bibitem[Heppert et~al.(2023)Heppert, Irshad, Zakharov, Liu, Ambrus, Bohg, Valada, and Kollar]{heppert2023carto}
Nick Heppert, Muhammad~Zubair Irshad, Sergey Zakharov, Katherine Liu, Rares~Andrei Ambrus, Jeannette Bohg, Abhinav Valada, and Thomas Kollar.
\newblock Carto: Category and joint agnostic reconstruction of articulated objects.
\newblock In \emph{Proceedings of the IEEE/CVF Conference on Computer Vision and Pattern Recognition}, pages 21201--21210, 2023.

\bibitem[Liu et~al.(2023{\natexlab{c}})Liu, Mahdavi-Amiri, and Savva]{liu2023paris}
Jiayi Liu, Ali Mahdavi-Amiri, and Manolis Savva.
\newblock Paris: Part-level reconstruction and motion analysis for articulated objects.
\newblock In \emph{Proceedings of the IEEE/CVF International Conference on Computer Vision}, pages 352--363, 2023{\natexlab{c}}.

\bibitem[Wang et~al.(2019)Wang, Zhou, Shi, Chen, Zhao, and Xu]{wang2019shape2motion}
Xiaogang Wang, Bin Zhou, Yahao Shi, Xiaowu Chen, Qinping Zhao, and Kai Xu.
\newblock Shape2motion: Joint analysis of motion parts and attributes from 3d shapes.
\newblock In \emph{Proceedings of the IEEE/CVF Conference on Computer Vision and Pattern Recognition}, pages 8876--8884, 2019.

\bibitem[Yan et~al.(2020)Yan, Hu, Yan, Chen, Van~Kaick, Zhang, and Huang]{yan2020rpm}
Zihao Yan, Ruizhen Hu, Xingguang Yan, Luanmin Chen, Oliver Van~Kaick, Hao Zhang, and Hui Huang.
\newblock Rpm-net: recurrent prediction of motion and parts from point cloud.
\newblock \emph{arXiv preprint arXiv:2006.14865}, 2020.

\bibitem[Weng et~al.(2021)Weng, Wang, Zhou, Qin, Duan, Fan, Chen, Su, and Guibas]{weng2021captra}
Yijia Weng, He~Wang, Qiang Zhou, Yuzhe Qin, Yueqi Duan, Qingnan Fan, Baoquan Chen, Hao Su, and Leonidas~J Guibas.
\newblock Captra: Category-level pose tracking for rigid and articulated objects from point clouds.
\newblock In \emph{Proceedings of the IEEE/CVF International Conference on Computer Vision}, pages 13209--13218, 2021.

\bibitem[Abdul-Rashid et~al.(2022)Abdul-Rashid, Freeman, Abbatematteo, Konidaris, and Ritchie]{abdul2022learning}
Hameed Abdul-Rashid, Miles Freeman, Ben Abbatematteo, George Konidaris, and Daniel Ritchie.
\newblock Learning to infer kinematic hierarchies for novel object instances.
\newblock In \emph{2022 International Conference on Robotics and Automation (ICRA)}, pages 8461--8467. IEEE, 2022.

\bibitem[Heiden et~al.(2022)Heiden, Liu, Vineet, Coumans, and Sukhatme]{heiden2022inferring}
Eric Heiden, Ziang Liu, Vibhav Vineet, Erwin Coumans, and Gaurav~S Sukhatme.
\newblock Inferring articulated rigid body dynamics from rgbd video.
\newblock In \emph{2022 IEEE/RSJ International Conference on Intelligent Robots and Systems (IROS)}, pages 8383--8390. IEEE, 2022.

\bibitem[Li et~al.(2021)Li, Yu, Sang, Wang, Song, Liu, Yeh, Zhu, Gundavarapu, Shi, Bi, Yu, Xu, Sunkavalli, Hasan, Ramamoorthi, and Chandraker]{li21openrooms}
Zhengqin Li, Ting{-}Wei Yu, Shen Sang, Sarah Wang, Meng Song, Yuhan Liu, Yu{-}Ying Yeh, Rui Zhu, Nitesh~B. Gundavarapu, Jia Shi, Sai Bi, Hong{-}Xing Yu, Zexiang Xu, Kalyan Sunkavalli, Milos Hasan, Ravi Ramamoorthi, and Manmohan Chandraker.
\newblock Openrooms: An open framework for photorealistic indoor scene datasets.
\newblock In \emph{{IEEE} Conference on Computer Vision and Pattern Recognition, {CVPR} 2021, virtual, June 19-25, 2021}, pages 7190--7199. Computer Vision Foundation / {IEEE}, 2021.
\newblock \doi{10.1109/CVPR46437.2021.00711}.
\newblock URL \url{https://openaccess.thecvf.com/content/CVPR2021/html/Li\_OpenRooms\_An\_Open\_Framework\_for\_Photorealistic\_Indoor\_Scene\_Datasets\_CVPR\_2021\_paper.html}.

\bibitem[Mao et~al.(2022)Mao, Zhang, Jiang, Chang, and Savva]{mao2022multiscan}
Yongsen Mao, Yiming Zhang, Hanxiao Jiang, Angel Chang, and Manolis Savva.
\newblock Multiscan: Scalable rgbd scanning for 3d environments with articulated objects.
\newblock \emph{Advances in Neural Information Processing Systems}, 35:\penalty0 9058--9071, 2022.

\bibitem[Deitke et~al.(2023{\natexlab{b}})Deitke, Hendrix, Farhadi, Ehsani, and Kembhavi]{deitke23phone2proc}
Matt Deitke, Rose Hendrix, Ali Farhadi, Kiana Ehsani, and Aniruddha Kembhavi.
\newblock Phone2proc: Bringing robust robots into our chaotic world.
\newblock In \emph{{IEEE/CVF} Conference on Computer Vision and Pattern Recognition, {CVPR} 2023, Vancouver, BC, Canada, June 17-24, 2023}, pages 9665--9675. {IEEE}, 2023{\natexlab{b}}.
\newblock \doi{10.1109/CVPR52729.2023.00932}.
\newblock URL \url{https://doi.org/10.1109/CVPR52729.2023.00932}.

\bibitem[Katz et~al.(2013)Katz, Kazemi, Bagnell, and Stentz]{katz2013interactive}
Dov Katz, Moslem Kazemi, J~Andrew Bagnell, and Anthony Stentz.
\newblock Interactive segmentation, tracking, and kinematic modeling of unknown 3d articulated objects.
\newblock In \emph{2013 IEEE International Conference on Robotics and Automation}, pages 5003--5010. IEEE, 2013.

\bibitem[{\"u}rgen Sturm et~al.(){\"u}rgen Sturm, Pradeep, Stachniss, Plagemann, Konolige, and Burgard]{urgenlearning}
J~{\"u}rgen Sturm, Vijay Pradeep, Cyrill Stachniss, Christian Plagemann, Kurt Konolige, and Wolfram Burgard.
\newblock Learning kinematic models for articulated objects.

\bibitem[Nie et~al.(2022)Nie, Gadre, Ehsani, and Song]{nie2022structure}
Neil Nie, Samir~Yitzhak Gadre, Kiana Ehsani, and Shuran Song.
\newblock Structure from action: Learning interactions for articulated object 3d structure discovery.
\newblock \emph{arXiv preprint arXiv:2207.08997}, 2022.

\bibitem[Jiang et~al.(2022)Jiang, Hsu, and Zhu]{jiang2022ditto}
Zhenyu Jiang, Cheng-Chun Hsu, and Yuke Zhu.
\newblock Ditto: Building digital twins of articulated objects from interaction.
\newblock In \emph{Conference on Computer Vision and Pattern Recognition (CVPR)}, 2022.

\bibitem[Ma et~al.(2023)Ma, Meng, Liu, Chen, Xu, and Chen]{ma2023sim2real}
Liqian Ma, Jiaojiao Meng, Shuntao Liu, Weihang Chen, Jing Xu, and Rui Chen.
\newblock Sim2real2: Actively building explicit physics model for precise articulated object manipulation.
\newblock In \emph{International Conference on Robotics and Automation (ICRA)}, 2023.

\bibitem[Hsu et~al.(2023)Hsu, Jiang, and Zhu]{hsu2023ditto}
Cheng-Chun Hsu, Zhenyu Jiang, and Yuke Zhu.
\newblock Ditto in the house: Building articulation models of indoor scenes through interactive perception.
\newblock \emph{arXiv preprint arXiv:2302.01295}, 2023.
\newblock URL \url{https://arxiv.org/abs/2302.01295}.

\bibitem[Memmel et~al.(2023{\natexlab{b}})Memmel, Wagenmaker, Zhu, Fox, and Gupta]{memmel2023asid}
Marius Memmel, Andrew Wagenmaker, Chuning Zhu, Dieter Fox, and Abhishek Gupta.
\newblock Asid: Active exploration for system identification and reconstruction in robotic manipulation.
\newblock In \emph{The Twelfth International Conference on Learning Representations}, 2023{\natexlab{b}}.

\bibitem[Ho et~al.(2020)Ho, Jain, and Abbeel]{ho2020denoising}
Jonathan Ho, Ajay Jain, and Pieter Abbeel.
\newblock Denoising diffusion probabilistic models.
\newblock \emph{Advances in neural information processing systems}, 33:\penalty0 6840--6851, 2020.
\newblock URL \url{https://arxiv.org/abs/2006.11239}.

\bibitem[Azizi et~al.(2023)Azizi, Kornblith, Saharia, Norouzi, and Fleet]{azizi2023synthetic}
Shekoofeh Azizi, Simon Kornblith, Chitwan Saharia, Mohammad Norouzi, and David~J Fleet.
\newblock Synthetic data from diffusion models improves imagenet classification.
\newblock \emph{arXiv preprint arXiv:2304.08466}, 2023.

\bibitem[Trabucco et~al.(2023{\natexlab{a}})Trabucco, Doherty, Gurinas, and Salakhutdinov]{trabucco2023effective}
Brandon Trabucco, Kyle Doherty, Max Gurinas, and Ruslan Salakhutdinov.
\newblock Effective data augmentation with diffusion models.
\newblock \emph{arXiv preprint arXiv:2302.07944}, 2023{\natexlab{a}}.

\bibitem[Li et~al.(2023)Li, Zhou, Zhang, Zhang, Wang, and Xie]{li2023open}
Ziyi Li, Qinye Zhou, Xiaoyun Zhang, Ya~Zhang, Yanfeng Wang, and Weidi Xie.
\newblock Open-vocabulary object segmentation with diffusion models.
\newblock In \emph{Proceedings of the IEEE/CVF International Conference on Computer Vision}, pages 7667--7676, 2023.

\bibitem[Yu et~al.(2023{\natexlab{a}})Yu, Li, Lou, Liu, Wan, Chen, and Li]{yu2023diffusion}
Xinyi Yu, Guanbin Li, Wei Lou, Siqi Liu, Xiang Wan, Yan Chen, and Haofeng Li.
\newblock Diffusion-based data augmentation for nuclei image segmentation.
\newblock In \emph{International Conference on Medical Image Computing and Computer-Assisted Intervention}, pages 592--602. Springer, 2023{\natexlab{a}}.

\bibitem[Wu et~al.(2023)Wu, Zhao, Chen, Gu, Zhao, He, Zhou, Shou, and Shen]{wu2023datasetdm}
Weijia Wu, Yuzhong Zhao, Hao Chen, Yuchao Gu, Rui Zhao, Yefei He, Hong Zhou, Mike~Zheng Shou, and Chunhua Shen.
\newblock Datasetdm: Synthesizing data with perception annotations using diffusion models.
\newblock \emph{arXiv preprint arXiv:2308.06160}, 2023.

\bibitem[Fu et~al.(2023)Fu, Tamir, Sundaram, Chai, Zhang, Dekel, and Isola]{fu23dreamsim}
Stephanie Fu, Netanel Tamir, Shobhita Sundaram, Lucy Chai, Richard Zhang, Tali Dekel, and Phillip Isola.
\newblock Dreamsim: Learning new dimensions of human visual similarity using synthetic data.
\newblock \emph{CoRR}, abs/2306.09344, 2023.
\newblock \doi{10.48550/arXiv.2306.09344}.
\newblock URL \url{https://doi.org/10.48550/arXiv.2306.09344}.

\bibitem[Tian et~al.(2023)Tian, Fan, Isola, Chang, and Krishnan]{tian23stablerep}
Yonglong Tian, Lijie Fan, Phillip Isola, Huiwen Chang, and Dilip Krishnan.
\newblock Stablerep: Synthetic images from text-to-image models make strong visual representation learners.
\newblock \emph{CoRR}, abs/2306.00984, 2023.
\newblock \doi{10.48550/arXiv.2306.00984}.
\newblock URL \url{https://doi.org/10.48550/arXiv.2306.00984}.

\bibitem[Jahanian et~al.(2022)Jahanian, Puig, Tian, and Isola]{jahanian22genmodels}
Ali Jahanian, Xavier Puig, Yonglong Tian, and Phillip Isola.
\newblock Generative models as a data source for multiview representation learning.
\newblock In \emph{The Tenth International Conference on Learning Representations, {ICLR} 2022, Virtual Event, April 25-29, 2022}. OpenReview.net, 2022.
\newblock URL \url{https://openreview.net/forum?id=qhAeZjs7dCL}.

\bibitem[Ho et~al.(2021)Ho, Rao, Xu, Jang, Khansari, and Bai]{ho2021retinagan}
Daniel Ho, Kanishka Rao, Zhuo Xu, Eric Jang, Mohi Khansari, and Yunfei Bai.
\newblock Retinagan: An object-aware approach to sim-to-real transfer.
\newblock In \emph{2021 IEEE International Conference on Robotics and Automation (ICRA)}, pages 10920--10926. IEEE, 2021.

\bibitem[Chen et~al.(2023{\natexlab{a}})Chen, Kiami, Gupta, and Kumar]{chen23genaug}
Zoey~Qiuyu Chen, Shosuke~C. Kiami, Abhishek Gupta, and Vikash Kumar.
\newblock Genaug: Retargeting behaviors to unseen situations via generative augmentation.
\newblock In Kostas~E. Bekris, Kris Hauser, Sylvia~L. Herbert, and Jingjin Yu, editors, \emph{Robotics: Science and Systems XIX, Daegu, Republic of Korea, July 10-14, 2023}, 2023{\natexlab{a}}.
\newblock \doi{10.15607/RSS.2023.XIX.010}.
\newblock URL \url{https://doi.org/10.15607/RSS.2023.XIX.010}.

\bibitem[Mandi et~al.(2022)Mandi, Bharadhwaj, Moens, Song, Rajeswaran, and Kumar]{mandi2022cacti}
Zhao Mandi, Homanga Bharadhwaj, Vincent Moens, Shuran Song, Aravind Rajeswaran, and Vikash Kumar.
\newblock Cacti: A framework for scalable multi-task multi-scene visual imitation learning.
\newblock \emph{arXiv preprint arXiv:2212.05711}, 2022.

\bibitem[Yu et~al.(2023{\natexlab{b}})Yu, Xiao, Tompson, Stone, Wang, Brohan, Singh, Tan, M, Peralta, Hausman, Ichter, and Xia]{yu23rosie}
Tianhe Yu, Ted Xiao, Jonathan Tompson, Austin Stone, Su~Wang, Anthony Brohan, Jaspiar Singh, Clayton Tan, Dee M, Jodilyn Peralta, Karol Hausman, Brian Ichter, and Fei Xia.
\newblock Scaling robot learning with semantically imagined experience.
\newblock In Kostas~E. Bekris, Kris Hauser, Sylvia~L. Herbert, and Jingjin Yu, editors, \emph{Robotics: Science and Systems XIX, Daegu, Republic of Korea, July 10-14, 2023}, 2023{\natexlab{b}}.
\newblock \doi{10.15607/RSS.2023.XIX.027}.
\newblock URL \url{https://doi.org/10.15607/RSS.2023.XIX.027}.

\bibitem[Bharadhwaj et~al.(2023)Bharadhwaj, Vakil, Sharma, Gupta, Tulsiani, and Kumar]{bharadhwaj2023roboagent}
Homanga Bharadhwaj, Jay Vakil, Mohit Sharma, Abhinav Gupta, Shubham Tulsiani, and Vikash Kumar.
\newblock Roboagent: Generalization and efficiency in robot manipulation via semantic augmentations and action chunking.
\newblock \emph{arXiv preprint arXiv:2309.01918}, 2023.

\bibitem[Trabucco et~al.(2023{\natexlab{b}})Trabucco, Doherty, Gurinas, and Salakhutdinov]{trabucco23dataaug}
Brandon Trabucco, Kyle Doherty, Max Gurinas, and Ruslan Salakhutdinov.
\newblock Effective data augmentation with diffusion models.
\newblock \emph{CoRR}, abs/2302.07944, 2023{\natexlab{b}}.
\newblock \doi{10.48550/arXiv.2302.07944}.
\newblock URL \url{https://doi.org/10.48550/arXiv.2302.07944}.

\bibitem[Zhang et~al.(2023)Zhang, Rao, and Agrawala]{zhang2023adding}
Lvmin Zhang, Anyi Rao, and Maneesh Agrawala.
\newblock Adding conditional control to text-to-image diffusion models.
\newblock In \emph{Proceedings of the IEEE/CVF International Conference on Computer Vision}, pages 3836--3847, 2023.
\newblock URL \url{https://arxiv.org/abs/2302.05543}.

\bibitem[Richardson et~al.(2023)Richardson, Metzer, Alaluf, Giryes, and Cohen-Or]{richardson2023texture}
Elad Richardson, Gal Metzer, Yuval Alaluf, Raja Giryes, and Daniel Cohen-Or.
\newblock Texture: Text-guided texturing of 3d shapes.
\newblock \emph{arXiv preprint arXiv:2302.01721}, 2023.
\newblock URL \url{https://arxiv.org/abs/2302.01721}.

\bibitem[Chen et~al.(2023{\natexlab{b}})Chen, Li, Lee, Tulyakov, and Nie{\ss}ner]{chen2023scenetex}
Dave~Zhenyu Chen, Haoxuan Li, Hsin-Ying Lee, Sergey Tulyakov, and Matthias Nie{\ss}ner.
\newblock Scenetex: High-quality texture synthesis for indoor scenes via diffusion priors.
\newblock \emph{arXiv preprint arXiv:2311.17261}, 2023{\natexlab{b}}.
\newblock URL \url{https://arxiv.org/abs/2311.17261}.

\end{thebibliography}

\end{document}